\documentclass[final,5p,times,twocolumn,longtitle]{elsarticle}

\usepackage{amssymb}
\usepackage{amsmath}

\usepackage{xcolor}
\usepackage{url}
\usepackage{multirow}
\usepackage{makecell}
\usepackage{booktabs}
\usepackage{tabularx}
\definecolor{todoColor}{rgb}{0.0,0.0,1.0}

\usepackage{xcolor}
\usepackage{booktabs,tabularx}
\usepackage{hyperref}
\hypersetup{
    colorlinks=true, 
    urlcolor=cyan,
    }
\usepackage{fontawesome5}

\usepackage{footmisc}

\newcommand{\teamtagmethod}[1]{\textcolor{#1}{\rule{1.2ex}{1.2ex}}\,}
\newcommand{\teamtagopen}[1]{\textcolor{#1}{\raisebox{0.2ex}{\scalebox{1.1}[0.8]{$\blacklozenge$}}}\,}
\newcommand{\teamtagsup}[1]{\textcolor{#1}{\raisebox{0.2ex}{\scalebox{0.9}{$\bigstar$}}}\,}

\definecolor{tcFOMO2JOMO}{HTML}{0072B2}  
\definecolor{tcDolphins}{HTML}{E69F00}   
\definecolor{tcTUKE}{HTML}{009E73}       
\definecolor{tcMaastricht}{HTML}{CC79A7} 
\definecolor{tcDKFZ}{HTML}{E7298A}       
\definecolor{tcMIPLAB}{HTML}{FF6A00}     
\definecolor{tcPrenuvo}{HTML}{6A3D9A}    
\definecolor{tcBrain}{HTML}{999999}      
\definecolor{tcAshash}{HTML}{32CD32}     
\definecolor{tcBiopet}{HTML}{008080}     

\definecolor{tcLatent}{HTML}{A6761D}     
\definecolor{tcHSEE}{HTML}{56B4E9}       
\definecolor{tcRaidium}{HTML}{7570B3}    
\definecolor{tcCAMI}{HTML}{66C2A5}       
\definecolor{tcMcGill}{HTML}{FC8D62}     
\definecolor{tcTiger}{HTML}{8DA0CB}      

\definecolor{tcSupIn}{HTML}{D55E00}      
\definecolor{tcSupOut}{HTML}{F0E442}     

\definecolor{placegold}{HTML}{D4AF37}
\definecolor{placesilver}{HTML}{A8A8A8}
\definecolor{placebronze}{HTML}{B87333}

\newcommand{\goldtrophy}{\textcolor{placegold}{\faTrophy}}
\newcommand{\silvertrophy}{\textcolor{placesilver}{\faTrophy}}
\newcommand{\bronzetrophy}{\textcolor{placebronze}{\faTrophy}}
  
\journal{Medical Image Analysis}

\begin{document}

\begin{frontmatter}



\title{Towards Brain MRI Foundation Models for the Clinic: Findings from the FOMO25 Challenge} 


\author[1,2]{Asbj{\o}rn Munk \corref{org}\fnref{eq}}\ead{asmu@di.ku.dk}
\author[1,2,3]{Stefano Cerri \corref{org}\fnref{eq}}\ead{stce@di.ku.dk}
\author[3,8]{Vardan Nersesjan\corref{org}}
\author[7]{Christian Hedeager Krag\corref{org}}
\author[1,2]{Jakob Ambsdorf\corref{org}}
\author[1,2]{Pablo Rocamora García\corref{org}}
\author[1,2]{Julia Machnio\corref{org}}
\author[6]{Peirong Liu\corref{org}}

\author[24]{Suhyun Ahn} 
\author[14]{Nasrin Akbari} 
\author[31]{Yasmina Al Khalil} 
\author[34,35,36]{Kimberly Amador} 
\author[44]{Sina Amirrajab} 
\author[33]{Tal Arbel} 
\author[29,25]{Meritxell Bach Cuadra} 
\author[37]{Ujjwal Baid} 
\author[37]{Bhakti Baheti} 
\author[25]{Jaume Banus} 
\author[43]{Kamil Barbierik} 
\author[17]{Christoph Brune} 
\author[23]{Yansong Bu} 
\author[41]{Baptiste Callard} 
\author[23]{Yuhan Chen} 
\author[33]{Cornelius Crijnen} 
\author[41]{Corentin Dancette} 
\author[19]{Peter Drotar} 
\author[38]{Prasad Dutande} 
\author[34,35,36]{Nils D. Forkert} 
\author[14]{Saurabh Garg} 
\author[21]{Jakub Gazda} 
\author[19]{Matej Gazda} 
\author[28]{Beno\^{i}t G\'{e}rin} 
\author[12]{Partha Ghosh} 
\author[22]{Weikang Gong} 
\author[29,25]{Pedro M. Gordaliza} 
\author[14]{Sam Hashemi} 
\author[18]{Tobias Heimann} 
\author[32]{Fucang Jia} 
\author[23]{Jiexin Jiang} 
\author[33]{Emily Kaczmarek} 
\author[34,35,36]{Chris Kang} 
\author[40]{Seung Kwan Kang} 
\author[34,35,36]{Mohammad Khazaei} 
\author[41]{Julien Khlaut} 
\author[30]{Petros Koutsouvelis} 
\author[40]{Jae Sung Lee} 
\author[32]{Yuchong Li} 
\author[23]{Mengye Lyu} 
\author[22]{Mingchen Ma} 
\author[42,37]{Anant Madabhushi} 
\author[12,48]{Klaus H. Maier-Hein} 
\author[41]{Pierre Manceron} 
\author[12]{Andrés Martínez Mora} 
\author[15]{Moona Mazher} 
\author[18]{Felix Meister} 
\author[25,27]{Nataliia Molchanova} 
\author[16]{Steven A Niederer} 
\author[45,46,30,47]{Leonard Nürnberg} 
\author[24]{Jinah Park} 
\author[16]{Abdul Qayyum} 
\author[29,25]{Jonas Richiardi} 
\author[41]{Antoine Saporta} 
\author[19]{Branislav Setlak} 
\author[13]{Ning Shen} 
\author[33]{Justin Szeto} 
\author[12]{Constantin Ulrich} 
\author[17]{Puru Vaish} 
\author[34,35,36]{Vibujithan Vigneshwaran} 
\author[30]{Leroy Volmer} 
\author[23]{Zihao Wang} 
\author[32]{Siqi Wei} 
\author[34,35]{Anthony Winder} 
\author[17]{Jelmer M. Wolterink} 
\author[26,28]{Maxence Wynen} 
\author[22]{Chang Yang} 
\author[39]{Si Young Yie} 
\author[1,2]{Mostafa Mehdipour Ghazi\corref{org}}
\author[11]{Akshay Pai\corref{org}}
\author[10]{Espen Jimenez Solem\corref{org}}
\author[1,2]{Sebastian N{\o}rgaard Llambias\corref{org}}
\author[7,9]{Mikael Boesen\corref{org}}
\author[3,10]{Michael Eriksen Benros\corref{org}}
\author[4,5]{Juan Eugenio Iglesias\corref{org}}
\author[1,2]{Mads Nielsen\corref{org}}

\affiliation[1]{organization={Department of Computer Science, University of Copenhagen},
                city={Copenhagen},
                country={Denmark}}
\affiliation[2]{organization={Pioneer Centre for AI},
                city={Copenhagen},
                country={Denmark}}
\affiliation[3]{organization={Copenhagen Research Centre for Biological and Precision Psychiatry, Mental Health Centre Copenhagen, Copenhagen University Hospital},
                region={Capital Region of Denmark},
                city={Copenhagen},
                country={Denmark}}
\affiliation[4]{organization={Athinoula A. Martinos Center for Biomedical Imaging, Massachusetts General Hospital and Harvard Medical School},
                city={Boston},
                state={Massachusetts},
                country={USA}}
\affiliation[5]{organization={Computer Science and Artificial Intelligence Laboratory, Massachusetts Institute of Technology},
                city={Boston},
                state={Massachusetts},
                country={USA}}
\affiliation[6]{organization={Johns Hopkins University},
                city={Baltimore},
                state={Maryland},
                country={USA}}
\affiliation[7]{organization={Radiological AI Testcenter (RAIT)},
                region={Capital Region of Denmark},
                city={Copenhagen},
                country={Denmark}}           
\affiliation[8]{organization={Copenhagen University Hospital, Rigshospitalet},
                region={Capital Region of Denmark},
                city={Copenhagen},
                country={Denmark}}
\affiliation[9]{organization={Copenhagen University Hospital, Bispebjerg \& Frederiksberg Hospital},
                region={Capital Region of Denmark},
                city={Copenhagen},
                country={Denmark}}
\affiliation[10]{organization={Department of Clinical Medicine, Faculty of Health and Medical Sciences, University of Copenhagen},
                 city={Copenhagen},
                 country={Denmark}}
\affiliation[11]{organization={Cerebriu},
                 city={Copenhagen},
                 country={Denmark}}
\affiliation[12]{organization={Division of Medical Image Computing, German Cancer Research Center (DKFZ)},
                 city={Heidelberg},
                 country={Germany}}
\affiliation[13]{organization={University of British Columbia},
                 city={Vancouver},
                 state={British Columbia},
                 country={Canada}}
\affiliation[14]{organization={Prenuvo},
                 city={Vancouver},
                 state={British Columbia},
                 country={Canada}}
\affiliation[15]{organization={Hawkes Institute, Department of Computer Science, University College London},
                 city={London},
                 country={United Kingdom}}
\affiliation[16]{organization={National Heart and Lung Institute, Faculty of Medicine, Imperial College London},
                 city={London},
                 country={United Kingdom}}
\affiliation[17]{organization={Department of Applied Mathematics, Technical Medical Centre, University of Twente},
                 city={Enschede},
                 country={Netherlands}}
\affiliation[18]{organization={Digital Technology and Innovation, Siemens Healthineers},
                 city={Erlangen},
                 country={Germany}}
\affiliation[19]{organization={IISLAB, Technical University of Košice},
                 city={Košice},
                 country={Slovakia}}
\affiliation[20]{organization={Siemens Healthcare s.r.o.},
                 city={Bratislava},
                 country={Slovakia}}
\affiliation[21]{organization={2nd Department of Internal Medicine, Pavol Jozef Safarik University and L Pasteur University Hospital},
                 city={Košice},
                 country={Slovakia}}
\affiliation[22]{organization={Fudan University},
                 city={Shanghai},
                 country={China}}
\affiliation[23]{organization={Shenzhen Technology University},
                 city={Shenzhen},
                 country={China}}
\affiliation[24]{organization={KAIST},
                 city={Daejeon},
                 country={South Korea}}
\affiliation[25]{organization={Department of Radiology, Lausanne University Hospital and University of Lausanne},
                 city={Lausanne},
                 country={Switzerland}}
\affiliation[26]{organization={Louvain Neuroinflammation Imaging Lab (NIL), Université Catholique de Louvain},
                 city={Brussels},
                 country={Belgium}}
\affiliation[27]{organization={University of Applied Sciences and Arts Western Switzerland (HES-SO Valais)},
                 city={Sion},
                 country={Switzerland}}
\affiliation[28]{organization={ICTEAM, Université Catholique de Louvain},
                 city={Louvain-la-Neuve},
                 country={Belgium}}
\affiliation[29]{organization={CIBM Center for Biomedical Imaging},
                 city={Lausanne},
                 country={Switzerland}}
\affiliation[30]{organization={Department of Radiation Oncology (Maastro), GROW Research Institute for Oncology and Reproduction, Maastricht University Medical Centre+},
                 city={Maastricht},
                 country={The Netherlands}}
\affiliation[31]{organization={Department of Biomedical Engineering, Medical Image Analysis, Eindhoven University of Technology},
                 city={Eindhoven},
                 country={The Netherlands}}
\affiliation[32]{organization={Shenzhen Institutes of Advanced Technology, Chinese Academy of Sciences},
                 city={Shenzhen},
                 country={China}}
\affiliation[33]{organization={McGill University and Mila - Quebec AI Institute},
                 city={Montreal},
                 country={Canada}}
\affiliation[34]{organization={Hotchkiss Brain Institute and Department of Radiology, University of Calgary},
                 city={Calgary},
                 state={Alberta},
                 country={Canada}} 
\affiliation[35]{organization={Department of Radiology, University of Calgary},
                 city={Calgary},
                 state={Alberta},
                 country={Canada}} 
\affiliation[36]{organization={Alberta Children's Hospital Research Institute, Department of Clinical Neuroscience, University of Calgary},
                 city={Calgary},
                 state={Alberta},
                 country={Canada}} 
\affiliation[37]{organization={The Wallace H. Coulter Department of Biomedical Engineering, Georgia Tech and Emory University},
                 city={Atlanta},
                 state={Georgia},
                 country={USA}}
\affiliation[38]{organization={SGGS College of Engineering and Technology},
                 city={Nanded},
                 state={Maharashtra},
                 country={India}}
\affiliation[39]{organization={Seoul National University},
                 city={Seoul},
                 country={South Korea}}
\affiliation[40]{organization={Brightonix Imaging},
                 city={Seoul},
                 country={South Korea}}
\affiliation[41]{organization={Raidium},
                 city={Paris},
                 country={France}}
\affiliation[42]{organization={Atlanta VA Health Care System},
                 city={Decatur},
                 state={Georgia},
                 country={USA}}
\affiliation[43]{organization={SWAI a.s.},
                 city={Prague},
                 country={Czech Republic}}
\affiliation[44]{organization={The D-Lab, Department of Precision Medicine, GROW Research Institute for Oncology and Reproduction, Maastricht University},
                 city={Maastricht},
                 country={The Netherlands}}
\affiliation[45]{organization={Artificial Intelligence in Medicine (AIM) Program, Mass General Brigham, Harvard Medical School},                       city={Boston},
                 state={Massachusetts},
                 country={USA}}
\affiliation[46]{organization={Radiology and Nuclear Medicine, CARIM \& GROW, Maastricht University}, city={Maastricht},
                 country={The Netherlands}}
\affiliation[47]{organization={ Department of Radiation Oncology, Dana-Farber Cancer Institute, Brigham and Women’s Hospital, Harvard Medical School}, city={Boston},
                 state={Massachusetts},
                 country={USA}}
\affiliation[48]{organization={Pattern Analysis and Learning Group, Heidelberg University Hospital}, city={Heidelberg},
                 country={Germany}}

\cortext[org]{Challenge organizers.}
\fntext[eq]{These authors contributed equally. Author order may be adjusted for individual use.}





\begin{abstract} 
Clinical deployment of automated brain MRI analysis faces a fundamental challenge: clinical data is heterogeneous and noisy, and high-quality labels are prohibitively costly to obtain. Self-supervised learning (SSL) can address this by leveraging the vast amounts of unlabeled data produced in clinical workflows to train robust \textit{foundation models} that adapt out-of-domain with minimal supervision. However, the development of foundation models for brain MRI has been limited by small pretraining datasets and in-domain benchmarking focused on high-quality, research-grade data. To address this gap, we organized the FOMO25 challenge as a satellite event at MICCAI 2025. FOMO25 provided participants with a large pretraining dataset, FOMO60K, and evaluated models on data sourced directly from clinical workflows in few-shot and out-of-domain settings. Tasks covered infarct classification, meningioma segmentation, and brain age regression, and considered both models trained on FOMO60K (Method Track) and any data (Open Track). Nineteen foundation models from sixteen teams were evaluated using a standardized containerized pipeline. Results show that (a) self-supervised pretraining improves generalization on clinical data under domain shift, with the strongest models trained \textit{out-of-domain} surpassing supervised baselines trained \textit{in-domain}. (b) No single pretraining objective benefits all tasks: MAE favors segmentation, hybrid reconstruction-contrastive objectives favor classification, and~(c) strong performance was achieved by small pretrained models, and improvements from scaling model size and training duration did not yield reliable benefits.

\end{abstract}

\end{frontmatter}


\section{Introduction}
The dominant paradigm in machine learning for neuroimaging is to train task-specific supervised models using manually annotated data. This paradigm has led to strong performance on internal validation sets and public benchmarks~\cite{Isensee2021,isensee2024}. However, performance often degrades when models are applied to real-world clinical data or to data acquired at new sites, vendors, protocols, or patient populations \cite{martensson2020,albadawy2018,llambias2025}. Models trained under the fully supervised paradigm are inherently constrained by the availability of labeled data, which is labor-intensive and requires specialist expertise to annotate. This creates a structural asymmetry: routine clinical workflows generate large amounts of Magnetic Resonance Imaging (MRI) scans~\cite{smith2012,smith2019}, but only a small fraction -- typically only for research purposes -- are labeled for supervised training.

The mismatch between abundant raw data and scarce labels is increasingly addressed through self-supervised learning (SSL) \cite{devlin2019,chen2020}. In this paradigm, the learning problem is fundamentally redefined: instead of training a separate predictor for each labeled task, one first trains an encoder on a large and diverse unlabeled corpus, before adapting the model to specific tasks. Evidence from natural language processing and computer vision suggests that such \textit{foundation models} are label-efficient, robust to distribution shift, and cheaper to finetune to new tasks \cite{Oquab2023,shi2023,He2022, brown2020language, liu2019roberta}.

In this light, large-scale structural MRI datasets present in clinical archives provide a compelling opportunity for pretraining foundation models in neuroimaging. However, progress in training foundation models for brain MRI has been limited by small pretraining datasets (<10K scans) and narrow evaluation \cite{tang2022,chen2023,wu2024}. Prior work has evaluated performance on clean, research-grade downstream benchmarks, where supervised models already perform well \cite{wald2025}. In this setting, foundation models are evaluated on their ability to provide cosmetic improvements over models which are already producing competitive segmentation masks, rather than benchmarking performance on tasks that reflect the noisy, heterogeneous reality of clinical practice.

To address this, we organized the Foundation Model Challenge for Brain MRI 2025 (FOMO25), held in conjunction with the 2025 International Conference on Medical Image Computing and Computer Assisted Intervention (MICCAI) in Daejeon, South Korea. FOMO25 aimed to advance SSL for brain MRI by (i) releasing FOMO60K~\cite{Cerri2026a}, a large-scale unlabeled dataset comprising 60,529 structural brain MRI scans from both clinical and research settings, and (ii) evaluating models on data sourced directly from clinical workflows. Submissions were evaluated in a challenging setting emphasizing out-of-domain generalization and label efficiency. Participants pretrained models in two tracks: a Method Track constrained to FOMO60K, and an Open Track with no restrictions on pretraining data, before finetuning them on three downstream tasks: infarct classification, meningioma segmentation, and brain age regression.

The challenge received 137 registrations before the deadline. From these, 30 participants representing 18 unique teams submitted a final foundation model for evaluation on the test set. In total, 19 unique foundation models were evaluated, each finetuned by the participants for the classification, segmentation, and regression tasks.

\begin{figure*}[!t]
    \centering
    \includegraphics[width=\linewidth]{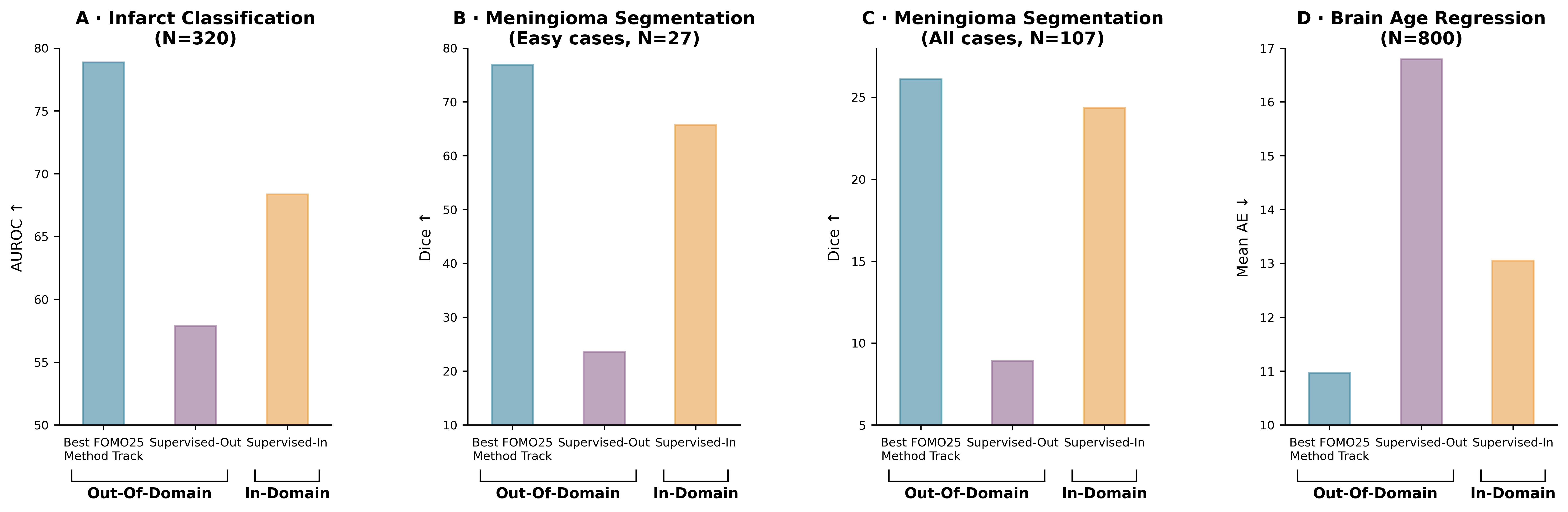}
    \caption{\textbf{Self-supervised pretraining boosts generalization.} Across tasks, the top pretraining-based model from the \textit{Method Track} outperforms both from-scratch out-of-domain and in-domain supervised baselines, demonstrating that SSL can effectively leverage heterogeneous MRI data. Baselines are nnU-Net (segmentation) and Asparagus (classification/regression). The best performing method for classification was \teamtagmethod{tcAshash}ashash, for segmentation \teamtagmethod{tcDolphins}Dolphin\_creators, and for regression \teamtagmethod{tcFOMO2JOMO}FOMO2JOMO. For Task 2, ``Easy cases'' were defined as the top 25\% of cases ranked by average Dice score across all challenge participants.}
    \label{fig:ssl_vs_sup}
\end{figure*}

The results of the FOMO25 challenge provide unique insights into the state of the art of training foundation models for neuroimaging and illustrate how design choices affect downstream performance\footnote{The FOMO25 challenge is reported in accordance with the BIAS reporting guideline \citep{Maier2020}.}. In particular, we find:

\begin{enumerate}
    \item Self-supervised pretraining boosts generalization: it not only outperforms supervised baselines trained on the same out-of-domain data, but the best out-of-domain foundation models even surpass specialist models trained in-domain (cf. Figure~\ref{fig:ssl_vs_sup}).
    \item No single SSL objective benefits classification, segmentation, and regression tasks equally: Local SSL objectives (e.g., Masked Autoencoders (MAE)~\cite{He2022}) tend to favor segmentation; hybrid SSL objectives (e.g., MAE + contrastive learning~\cite{chen2020}) tend to favor classification; while all objective types seem to perform equally well on regression (cf. Figure~\ref{fig:ptstrategy} (A) - (C)).
    \item Small models were competitive. Larger model sizes and longer pretraining did not yield meaningful gains on downstream performance (cf. Figure~\ref{fig:ptstrategy} (D) - (E)).
\end{enumerate}

\subsection{Related Work}

Given the recency of the field, only a few challenges have systematically benchmarked self-supervised learning and foundation models in medical imaging. Concurrently with FOMO25, the Self-Supervised Learning for 3D Medical Imaging (SSL3D) Challenge~\cite{SSL3D2025} was held in conjunction with MICCAI 2025. SSL3D used a large 3D MRI dataset (114,570 volumes from 34,191 subjects)~\cite{Wald2025b} and standardized finetuning and evaluation protocols to compare SSL strategies across segmentation and classification tasks. However, the challenge restricted participants to two backbone architectures (a 3D CNN and a ViT), both with approximately 100M parameters, which made central finetuning by the challenge organizers possible. This enabled a controlled assessment of pretraining objectives on volumetric data, but analysis was constrained to the provided backbones, preprocessing pipeline, and finetuning recipe, and did not allow for analysis on how alternative strategies affect transfer. In contrast, the FOMO25 challenge did not enforce such constraints on submissions, allowing for a broader investigation of how pretraining influences downstream adaptation, at the cost of reduced disentanglement of the pretext task. Further, while SSL3D evaluates methods on clinical data, the evaluation is largely focused on in-domain and well-annotated settings, while FOMO25 is mainly focused on few-shot and label-sparse scenarios.

Another recent effort, the CVPR 2025 “Foundation Models for Interactive 3D Biomedical Image Segmentation” challenge~\cite{CVPR2025}, explored how foundation models could be trained specifically for 3D segmentation to adapt across modalities and anatomical targets from interactive prompts, using a large set of 3D image-mask pairs. While the challenge focused on interactive segmentation rather than few‑shot adaptation and included CT, MRI, and PET imaging from diverse body regions, it highlighted the utility of large-scale pretraining and standardized evaluation frameworks across diverse biomedical imaging contexts. However, FOMO25 differed fundamentally by treating foundation models as a general representation learning tool that should be able to effectively adapt to diverse tasks, including classification, regression, and segmentation. Further, FOMO25 evaluated representations by finetuning, not prompting, with a clear focus on evaluation on data sourced from clinical workflows.

\begin{table*}[ht]
\centering
\small
\setlength{\tabcolsep}{10pt}
\renewcommand{\arraystretch}{1.15}
\begin{tabular}{
  r l c
  @{\hspace{14pt}} cc
  @{\hspace{10pt}} ccc
  @{\hspace{10pt}} ccc
}
\toprule
& & \multicolumn{1}{c}{Overall} &
\multicolumn{2}{c}{Task 1 (Infarct)} &
\multicolumn{3}{c}{Task 2 (Meningioma)} &
\multicolumn{3}{c}{Task 3 (Brain age)} \\
\cmidrule(lr){3-3}
\cmidrule(lr){4-5}
\cmidrule(lr){6-8}
\cmidrule(lr){9-11}
Place & Team &
\textbf{Rank} &
\textbf{R} & \textbf{AUROC} &
\textbf{R} & \textbf{Dice} & \textbf{NSD} &
\textbf{R} & \textbf{Corr.} & \textbf{Mean AE} \\
\midrule

\multicolumn{11}{c}{\textbf{Method Track}}\\
\midrule
\goldtrophy\;1   & \teamtagmethod{tcFOMO2JOMO}FOMO2JOMO           & 2.3  & 2  & 0.782 & 4  & 0.095 & 0.072 & 1 & 0.651 & 10.96 \\
\silvertrophy\;2 & \teamtagmethod{tcDolphins}Dolphins\_creators   & 4.3  & 6  & 0.684 & 1  & 0.261 & 0.232 & 6 & 0.448 & 12.67 \\
\bronzetrophy\;3 & \teamtagmethod{tcMaastricht}MaastrichtU-CDS    & 4.7  & 5  & 0.688 & 2.5 & 0.181 & 0.126 & 6.5 & 0.599 & 16.17 \\
4                  & \teamtagmethod{tcTUKE}TUKE                   & 4.8  & 4  & 0.750 & 7  & 0.046 & 0.043 & 3.5 & 0.602 & 11.96 \\ 
5                  & \teamtagmethod{tcMIPLAB}MIPLAB               & 5.3  & 9  & 0.473 & 5  & 0.065 & 0.053 & 2 & 0.632 & 11.48 \\
6                  & \teamtagmethod{tcDKFZ}DKFZ\_MIC              & 5.5  & 8  & 0.569 & 2.5  & 0.171 & 0.155 & 6 & 0.557 & 13.31 \\
6                  & \teamtagmethod{tcPrenuvo}prenuvo             & 5.5  & 7  & 0.649 & 6  & 0.064 & 0.047 & 3.5 & 0.626 & 12.53 \\
8                  & \teamtagmethod{tcBrain}fomo25\_brain         & 6.2  & 3  & 0.760 & 8  & 0.006 & 0.005 & 7.5 & 0.409 & 14.85 \\
9                  & \teamtagmethod{tcAshash}ashash               & 6.5  & 1  & 0.789 & 9.5 & 0.003 & 0.004 & 9 & 0.370 & 17.62 \\
10                 & \teamtagmethod{tcBiopet}biopet               & 9.8  & 10 & 0.440 & 9.5 & 0.005 & 0.002 & 10& 0.137 & 18.21 \\


\midrule
\multicolumn{11}{c}{\textbf{Open Track}}\\
\midrule
\goldtrophy\;1  & \teamtagopen{tcLatent}LatentCampus          & 2.7  & 2 & 0.659 & 1 & 0.275 & 0.243 & 5 & 0.558  & 16.50 \\
\silvertrophy\;2 & \teamtagopen{tcFOMO2JOMO}FOMO2JOMO          & 3.7  & 4 & 0.625 & 6 & 0.039 & 0.040 & 1 & 0.861  & 6.80  \\
\bronzetrophy\;3 & \teamtagopen{tcDKFZ}DKFZ\_MIC               & 4.2 & 7 & 0.569 & 2 & 0.252 & 0.227 & 3.5 & 0.556  & 13.67 \\
4                  & \teamtagopen{tcMIPLAB}MIPLAB                & 4.5 & 3 & 0.639 & 5 & 0.048 & 0.046 & 5.5 & 0.320  & 14.21 \\
5                  & \teamtagopen{tcHSEE}FOMO HSEE               & 4.7  & 9 & 0.491 & 3 & 0.184 & 0.141 & 2 & 0.633  & 12.27 \\
6                  & \teamtagopen{tcRaidium}Curia (Raidium)     & 4.8  & 5 & 0.593 & 4 & 0.110 & 0.093 & 5.5 & 0.357  & 14.31 \\
7                  & \teamtagopen{tcCAMI}CAMI\_SIAT              & 5.5  & 1 & 0.671 & 8.5 & 0.007 & 0.004 & 7 & 0.227  & 16.47 \\
8                  & \teamtagopen{tcMcGill}McGill\_PVG           & 7.3  & 8 & 0.560 & 7.5 & 0.015 & 0.007 & 7 & 0.444  & 18.03 \\
9                  & \teamtagopen{tcTiger}TeamTigerAtFOMO2025    & 7.5  & 6 & 0.580 & 8 & 0.006 & 0.007 & 8.5 & -0.032 & 16.79 \\
\bottomrule
\end{tabular}

\caption{FOMO25 leaderboard results.}
\label{tab:results}
\end{table*}

\section{Data}

This section summarizes the datasets used in the FOMO25 challenge. We first describe the pretraining corpus for the \emph{Method Track} (FOMO60K). We then present the downstream datasets for the three challenge tasks --- infarct classification (Task 1), meningioma segmentation (Task 2), and brain-age regression (Task 3) --- along with the access model, preprocessing, and labeling protocols.

\subsection{Pretraining data}
\label{subsec:pretraining}
For the \emph{Method Track}, participants were required to use FOMO60K~\cite{Cerri2026a} for model pretraining. FOMO60K comprises 60,529 MRI scans from 13,900 sessions across 11,187 subjects, aggregated from 16 publicly available sources, and was released in conjunction with this challenge on Huggingface\footnote{\url{https://huggingface.co/datasets/FOMO-MRI/FOMO60K}}. The dataset is highly heterogeneous in terms of subject demographics, scanner vendors and models, field strengths, acquisition protocols, and MRI sequence types. All images were reoriented to RAS orientation and affinely co-registered within each session to the highest-resolution scan. Each scan was either skull-stripped or defaced. Further details are provided in~\cite{Cerri2026a}.

\subsection{Finetuning datasets}
For finetuning, the classification and segmentation datasets were provided by the Copenhagen startup Cerebriu, originating from multiple hospitals in India, while Task 3 was drawn from a subset of the dataset in~\cite{Iglesias2023} acquired in Boston, USA. The datasets were intentionally kept small to evaluate the few-shot learning capabilities of pretrained models in realistic data-limited settings. For the remainder of the paper, we define a \textit{case} as the set of MRI sequences provided for a single subject at a single imaging session, together with the associated task-specific labels.

\begin{itemize}
    \item \textbf{Task 1}: 21 cases were used for finetuning, including 13 with infarcts (positive class) and 8 without (negative class). Each case included FLAIR, DWI (ADC and b=1000), and either T2* or SWI images. V.N. and C.H.K. provided infarct labels.
    
    \item \textbf{Task 2}: 23 cases were used for finetuning. Scans included FLAIR, DWI (b=1000), and either T2* or SWI images, accompanied by a binary mask delineating the meningiomas. Masks were manually annotated by V.N. and C.H.K. following a standardized protocol.
    
    \item \textbf{Task 3}: 200 cases were included for finetuning, each consisting of a T1w and a T2w scan. Labels corresponded to the subject's chronological age at the time of the scan, in years.
\end{itemize}

\noindent
No patient-level metadata beyond the task labels was provided to participants.
Additional details on labeling protocols, MRI sequences, and subject demographics are provided in~\ref{app:Datasets}.

\subsection{Validation and testing datasets}

Validation and test datasets were acquired from multiple hospitals in Denmark~\cite{Cerri2026b}, with MRI sequences and labeling protocols identical to those used for finetuning, but originating from different institutions and geographic regions, thereby constituting an out-of-domain evaluation setting. The data were split approximately 20\% for validation and 80\% for testing. Specifically, Task 1 included 80 validation and 320 test scans, with equal numbers of positive and negative cases; Task 2 included 25 validation and 107 test scans; and Task 3 included 200 validation and 800 test scans. Additional details on MRI sequences, labeling procedures, and subject demographics are provided in~\ref{app:Datasets}.

\subsection{Data availability and licensing}

The FOMO60K pretraining dataset is released under the CC BY-NC-SA 4.0 license, with individual constituent datasets retaining their original licenses as described in~\cite{Cerri2026a}. Finetuning data for Tasks 1 and 2 is restricted to use within this challenge. Task 3 finetuning data follows the terms of~\cite{Iglesias2023}. Validation and test data remain private.

\section{Design of the FOMO25 Challenge}
\subsection{Mission}
FOMO25 was designed to provide a clinically grounded benchmark for brain MRI foundation models that moves beyond marginal gains on research datasets. The mission was to quantify how self-supervised pretraining on heterogeneous, real-world MRI transfers to multiple downstream problems under realistic constraints: (i) \emph{few-shot} labeled finetuning, reflecting limited annotation capacity in practice, and (ii) \emph{out-of-domain} evaluation across sites and vendors, reflecting deployment conditions. To go beyond well-studied supervised baselines, the challenge tasks span complementary objectives (classification, segmentation, and regression) and are evaluated on clinical-grade, multi-center data. Methodological constraints were kept minimal to encourage diverse pretraining paradigms and architectures, while standardizing the finetuning data and evaluation protocol to avoid dataset-specific hacks and ensure that leaderboard differences primarily reflect representation quality and transfer robustness. 

\subsection{Tracks and rules}

\begin{figure*}[!t]
    \centering
    \includegraphics[width=0.95\linewidth]{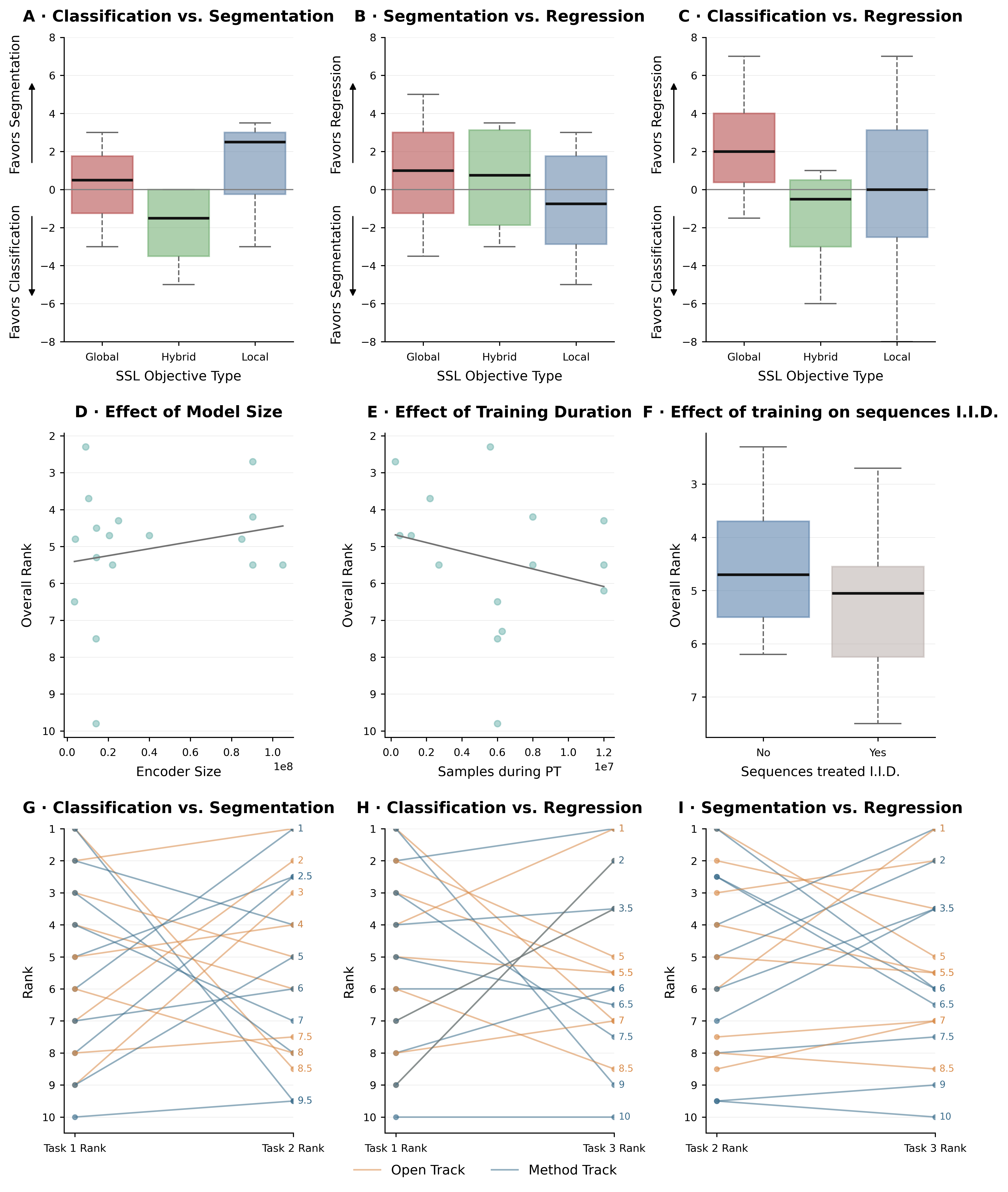}
    \caption{
    \textbf{Effect of pretraining choices on downstream performance.} (A--C) Pairwise rank differences between tasks, grouped by SSL objective category (global, hybrid, local): classification vs.\ segmentation~(A), segmentation vs.\ regression~(B), and classification vs.\ regression~(C). Teams with Dice or NSD < 0.01 were excluded from (A) and (B). Positive values indicate better relative performance on the task shown at the top of the axis. (D) Overall rank as a function of encoder size. One outlier with 0.5B params was removed from the plot.
    (E) Overall rank as a function of the number of samples seen during pretraining. (F) Overall rank by whether sequences were treated as i.i.d.\ during pretraining. (G--I) Per-method rank comparisons across task pairs: classification vs.\ segmentation~(G), classification vs.\ regression~(H), and segmentation vs.\ regression~(I), with lines colored by track. 
    }
    \label{fig:ptstrategy}
\end{figure*}

\begin{figure*}[!t]
    \centering
    \includegraphics[width=\textwidth]{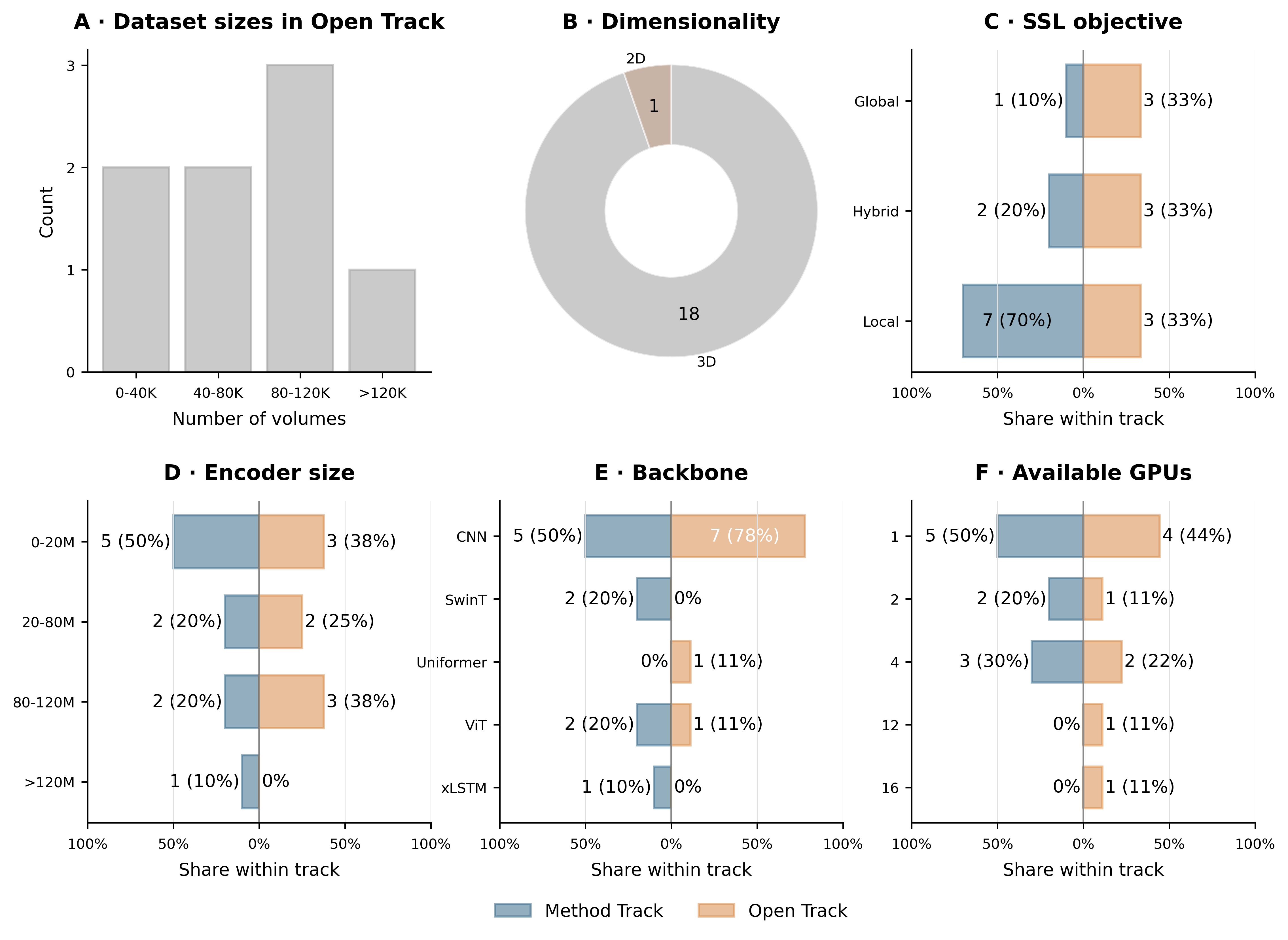}
    \caption{
    \textbf{Comparison of Method and Open Track Submissions.}
    (A) Distribution of pretraining dataset sizes in the Open Track (Method Track teams all used FOMO60K). (B) Dimensionality of the input representation (2D vs.\ 3D). (C--F) Share of submissions within each track by SSL objective category (global, hybrid, local)~(C), encoder size~(D), backbone architecture~(E), and number of available GPUs~(F)
    }
    \label{fig:track_comparison}
\end{figure*}

FOMO25 comprised two tracks designed to disentangle algorithmic advances from possible data scale effects:

\begin{itemize}
    \item \textbf{Method Track:} Participants pretrained their models exclusively on FOMO60K (see Section~\ref{subsec:pretraining} for more details). This constraint isolates methodological contributions by controlling for pretraining data, enabling comparison of architectures, pretraining objectives, and finetuning strategies across teams.
    \item \textbf{Open Track:} Participants were allowed to use any pretraining data, including private datasets, without restrictions. This track benchmarks the performance achievable with arbitrary data resources and provides insights into the practical benefits of large-scale or proprietary pretraining corpora.
\end{itemize} 

Both tracks used identical finetuning datasets and were evaluated on the same held-out out-of-domain validation and test datasets. Additional labeled finetuning data was prohibited in both tracks. No further restrictions were put on finetuning, since finetuning foundation models for brain MRI remains underexplored, and it is currently unclear which adaptation strategies are best suited to which pretrained models, backbones, etc.

\subsection{Submission and evaluation}

The following section provides a brief overview of how submissions were handled and evaluated.

\subsubsection{Infrastructure}
The challenge infrastructure comprised (i) code to reduce the workload required for participation, (ii) a set of public web resources for communication, data and code distribution, and submission management, and (iii) a secure compute environment for evaluation on sensitive data. 

Code provided to the participants covered a baseline codebase and a container validator. The baseline codebase\footnote{\url{https://github.com/fomo25/baseline-codebase}} provided a starting point for preprocessing, pretraining, and finetuning for each of the three tasks, as well as containerization of submissions. Provided pretraining code was based on the AMAES \cite{munk2024amaes} SSL methodology and Yucca \cite{llambias2024yucca} framework. The container validator\footnote{\url{https://github.com/fomo25/container-validator}} allowed participants to verify that their container was correctly configured and could be evaluated before submissions. 

The public web resources were:
\begin{enumerate}
    \item \href{https://fomo25.github.io}{fomo25.github.io}: The main challenge website, hosting the rules, timelines, and frequently asked questions, and providing navigation to the remaining resources.
    \item \href{https://synapse.org/Synapse:syn64895667}{synapse.org}: The submission portal used to manage the evaluation queue and to host the validation leaderboard.
    \item \href{https://huggingface.co/datasets/FOMO-MRI/FOMO60K}{huggingface.co}: Distribution of the FOMO60K pretraining dataset.
    \item \href{https://github.com/orgs/fomo25/repositories}{github.com}: Distribution of baseline code and the container validator used for local technical checks.
\end{enumerate}

\noindent
Since the validation and test datasets contain sensitive information, submitted methods were executed on a secure cluster hosted at the National Genome Center of Denmark\footnote{\href{https://www.eng.ngc.dk}{eng.ngc.dk}}, which is physically located in Copenhagen. The cluster is ISO-27001 certified and does not have general internet access, except for a controlled channel used to transfer submissions for evaluation.

\subsubsection{Submission of solutions}
\begin{figure*}[!t]
    \centering
    \includegraphics[width=\linewidth]{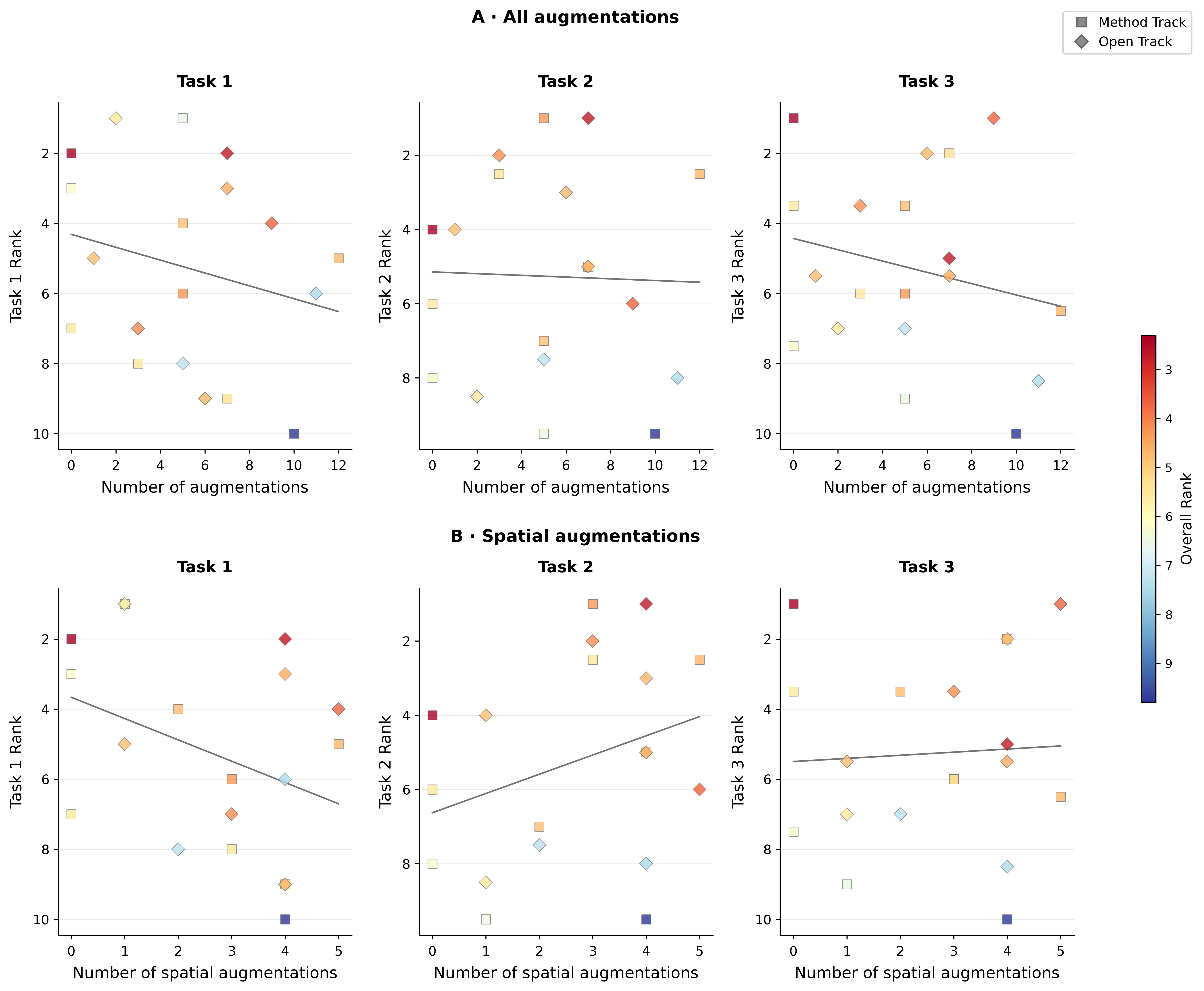}
    \caption{\textbf{Effect of augmentation strategies} on task performance. (A) Task-specific rank as a function of the total number of pretraining augmentations. (B) Task-specific rank as a function of the number of spatial augmentations only. Marker shapes denote track, and colors indicate overall rank within each track. Lower rank indicates better performance. Trend lines are shown for each task.}
    \label{fig:augmentation}
\end{figure*}

Participants first pretrained a foundation model and then finetuned it independently for each downstream task, resulting in three task-specific models derived from a common pretrained checkpoint (one each for infarct classification, meningioma segmentation, and brain-age regression). Each solution was submitted as an \emph{inference-only} \emph{Apptainer} container (\texttt{.sif}), i.e., the container implemented a standardized prediction interface and did not include any pretraining or finetuning code. Consequently, each team submitted three separate containers per track (one per task).

Before submission, participants could locally verify container build and input/output conventions using the provided container validator prior to uploading to Synapse. Submissions were handled via two Synapse-managed queues: a \emph{validation} queue with no limit on the number of submissions, and a \emph{test} queue used for the final evaluation, where only one submission per track, task, and team, was permitted. Containers in both queues were evaluated manually by the organizers by executing the submitted containers, collecting predictions, and computing the corresponding metrics for leaderboard reporting; operationally, most submitted containers were processed within approximately six hours of submission.

\subsubsection{Timeline}
The challenge opened for registration on 1 April 2025. The baseline pretraining and finetuning codebase was released on 7 April, followed by the release of the container sanity-check/validation tooling on 8 May. The submission pipeline and validation leaderboard opened on 15 June, enabling iterative validation submissions throughout the development period. The final submission deadline was extended from 20 August to 31 August 2025 (11:59 p.m. PST), after which teams selected for presentations were contacted on 2 September. The challenge concluded with the presentation of the final results at the MICCAI 2025 half-day event on 27 September 2025.

\subsection{Organization}
FOMO25 was organized as the first edition of a recurring benchmark intended to track progress in brain MRI foundation models over time. A follow-up challenge (FOMO26) has been proposed with the goal of maintaining a continuous series. Results and the full leaderboard were announced at the MICCAI 2025 half-day challenge event. The challenge offered a total cash prize of \$2000, with \$1000 allocated per track and distributed among the top-performing teams (the exact split across ranks was announced closer to the deadline). Winners were determined from the unified leaderboard within each track.

To manage conflicts of interest, the participation policy distinguished two groups affiliated with the organizers' institutes. First, participants with active working relations with any organizer were not permitted to participate and were not listed on the leaderboard. Second, other members of the organizers' institutes were allowed to participate, but were ineligible for awards and were listed on the leaderboard with a clearly visible disclaimer.

Teams that made substantial submissions (e.g., excluding trivial submissions such as all-zero outputs or random guessing) to all three tasks were invited to co-author this paper, with up to five authors per team and case-by-case exemptions. Participating teams retained the right to publish their own methods independently; an embargo until 15 November 2025 was offered to teams wishing to publish ahead of this paper. Participating teams were encouraged but not required to release their code.

\subsection{Metrics}
Following the recommendations of~\cite{Maier2024}, Task 1 was evaluated using the area under the receiver operating characteristic curve (AUROC) and Task 2 using the Dice similarity coefficient (Dice) and normalized surface distance (NSD). No standardized recommendations exist for Task 3, so we evaluated it using mean absolute error (Mean AE) and the Pearson correlation coefficient (Corr.). These metrics provide complementary perspectives: Mean AE quantifies absolute prediction accuracy, while correlation captures alignment with the ground truth and penalizes systematic bias. For test cases where a submission failed to produce a valid prediction, the worst possible metric value was assigned.

\section{Results}

Across the two tracks, 21 final submissions were received from 18 teams, with three teams submitting to both tracks. 16 teams submitted full, valid submissions covering all three tasks, while one team submitted to a subset of tasks, and another team's submission was invalid. Participation was active throughout the development phase: 322 containers were submitted to the validation queue. After the final submission, teams described their method by filling out a detailed questionnaire, ensuring all data was collected in a systematic manner which allowed for systematic analysis. The final leaderboard is given in Table \ref{tab:results}. We provide an overview of all submitted methods in Section \ref{sec:submissions_overview}. 

\subsection{Baselines}

For each task, we provide a fully supervised baseline to serve as a reference point for how well supervised models perform versus our two leaderboards. We distinguish two settings: \teamtagsup{tcSupOut}Supervised-OOD, where the model is trained on the out-of-domain finetune dataset, thus training on the same data as the challenge participants, and \teamtagsup{tcSupIn}Supervised-ID, where it is trained on the validation dataset, which is in-domain and notably larger than the finetune dataset. The baseline comparison and analysis is provided in Figure \ref{fig:stats_rankings}.

For the classification and regression tasks, we use the Asparagus framework~\cite{Cerri2026a}, as there is no established consensus on a fully supervised state-of-the-art method for these task types. The Asparagus models used a ResEnc\cite{isensee2024} model with 90M parameters, and a small classification head consisting of global average pooling and a linear layer, trained with SGD. For the segmentation task, we use nnU-Net~\cite{Isensee2021}, using the default planner, as this has consistently shown to achieve near-SOTA segmentation performance \cite{isensee2024}. We evaluated both the ensemble and full-dataset training configurations of nnU-Net and reported results for the best-performing variant on the test set, which was trained on the full dataset.

\subsection{Rankings and Statistical Analysis}

Submissions were ranked by first averaging metric-specific ranks within each task, and then across tasks to obtain an overall ranking. Ranking stability was assessed using 10,000 bootstrap resamples of test cases, recomputing ranks to derive 95\% confidence intervals. Statistical significance between submissions was evaluated using pairwise two-sided permutation testing (100,000 permutations) with Phipson-Smyth correction~\cite{Phipson2010}. For each pair of submissions, predictions were randomly swapped on a subject-by-subject basis, and task-specific test statistics (rank differences) were recomputed. Per-task p-values were combined using the maximum, ensuring that a team could only be considered superior if it outperformed the other across all tasks. Performance tiers ($\alpha = 0.05$) were assigned by sequentially grouping teams (ordered by mean rank) that did not show significant pairwise differences.

The bootstrap rank distributions and performance tiers for the combined ranking are shown in Figure~\ref{fig:stats_rankings} (A–B), with per-task breakdowns and pairwise permutation-test p-value matrices provided in Figure~\ref{fig:method_stats_analysis} and Figure~\ref{fig:open_stats_analysis}. In the Method Track, five performance tiers emerge, with \teamtagmethod{tcFOMO2JOMO}FOMO2JOMO alone in Tier 1, followed by five methods — including the \teamtagsup{tcSupIn}Supervised-ID baseline — grouped in Tier 2. The Open Track yields only three tiers, with \teamtagsup{tcSupIn}Supervised-ID alone in Tier 1 and \teamtagopen{tcLatent}LatentCampus in Tier 2, indicating that no open-track submission significantly outperformed the in-domain supervised baseline overall. Per-task rankings shift considerably: in the Method Track, the top-ranked team differs for each task, and no team remains in Tier 1 across all three. The classification task shows the fewest distinguishable tiers in both tracks (four in the Method Track, two in the Open Track), whereas Task 3 consistently yields the most tiers, with the narrowest confidence intervals and the most significant pairwise differences in the p-value matrices. Overall, the permutation-test p-value matrices confirm that significant pairwise differences are most concentrated between the extremes of each leaderboard, reinforcing that rankings should be interpreted in terms of tiers rather than exact ordinal positions \cite{maier2018rankings}.

\subsection{Effect of pretraining}
\begin{figure*}[!ht]
    \centering
    \includegraphics[width=\linewidth]{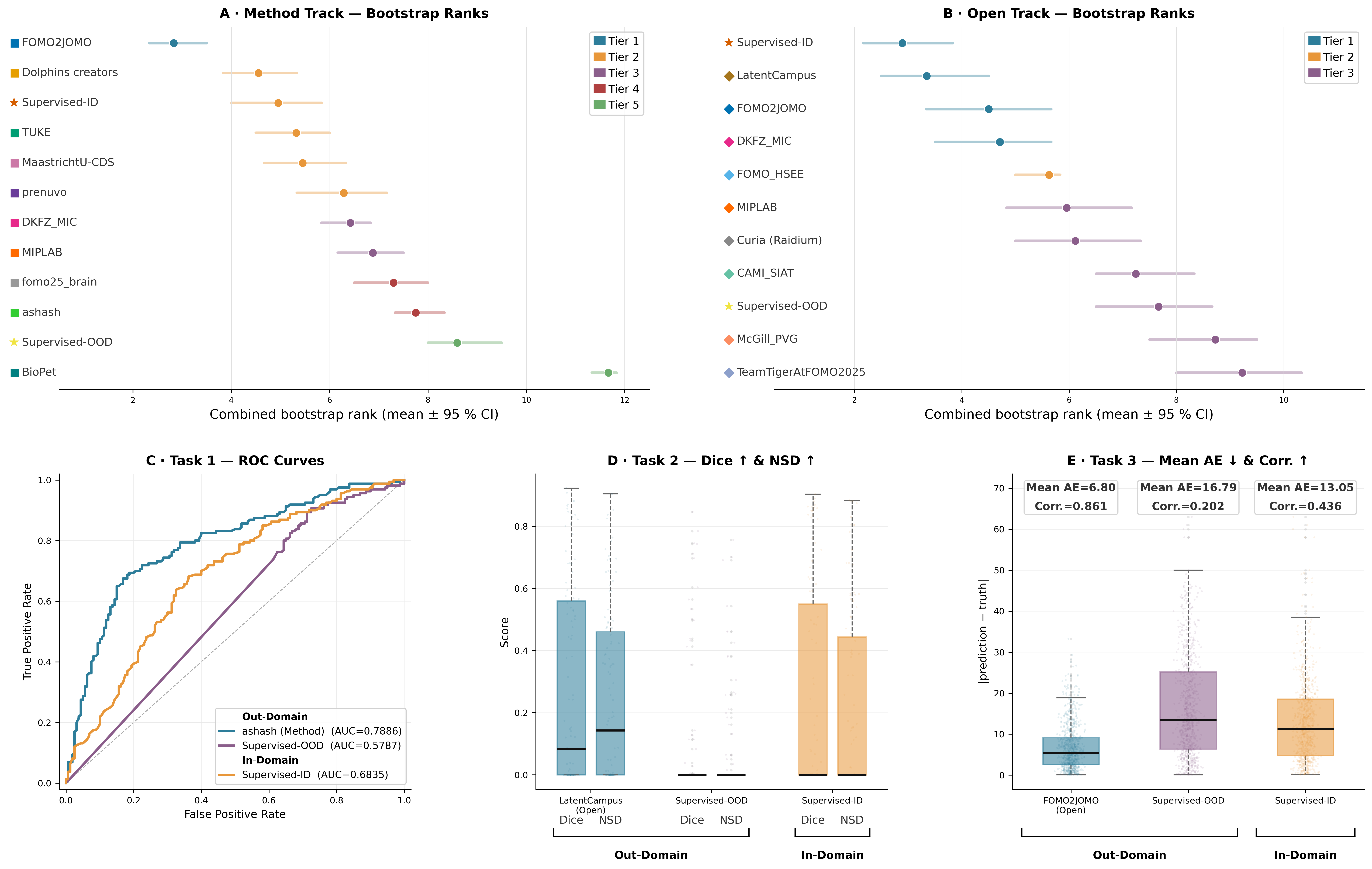}
    \caption{\textbf{Rankings with supervised baseline models.} (A, B) Bootstrap rank distributions (mean $\pm$ 95\% CI) for the Method and Open Tracks, with teams grouped into statistically defined performance tiers (colours). (C-E) Task-level performance for the best out-of-domain method across both tracks alongside in-domain and out-of-domain supervised baselines: (C) ROC curves for the infarct classification task, (D) Dice and NSD score distributions for the meningioma segmentation task, (E) absolute error distributions for the brain-age regression task, with Mean AE and Corr. annotated. Rankings per individual task, along with permutation-test p-value matrices, are provided in Figure~\ref{fig:method_stats_analysis} for the Method Track and in Figure~\ref{fig:open_stats_analysis} for the Open Track.}
    \label{fig:stats_rankings}
\end{figure*}

We provide a comparison between pretrained models and the supervised baselines in Figure \ref{fig:stats_rankings}. We note that 16 out of 19 models are ranked higher than \teamtagsup{tcSupOut}Supervised-OOD, which is trained on the same data, showing that pretrained foundation models in general provide measurable improvements over supervised learning. Even more importantly, two models \teamtagmethod{tcFOMO2JOMO}FOMO2JOMO and \teamtagmethod{tcDolphins}Dolphin\_creators beat \teamtagsup{tcSupIn}Supervised-ID, with the +2 overall rank improvement of \teamtagmethod{tcFOMO2JOMO}FOMO2JOMO being statistically significant.

Further evidence for the benefit of pretraining is seen when comparing the best-performing model from the Method Track on each task against the supervised baselines, as given in Figure \ref{fig:ssl_vs_sup}. On the classification task, \teamtagmethod{tcAshash}ashash improves performance by $20.99$ AUC points over \teamtagsup{tcSupOut}Supervised-OOD and by $10.51$ AUC points over \teamtagsup{tcSupIn}Supervised-ID. On Task 2, \teamtagopen{tcLatent}Dolphin\_creators achieves a gain of $17.20$ Dice points relative to \teamtagsup{tcSupOut}Supervised-OOD and $1.75$ Dice points relative to \teamtagsup{tcSupIn}Supervised-ID. On Task 3, \teamtagmethod{tcFOMO2JOMO}FOMO2JOMO reduces the error by $5.83$ Mean AE points compared with \teamtagsup{tcSupOut}Supervised-OOD and by $2.09$ Mean AE points compared with \teamtagsup{tcSupIn}Supervised-ID. 

\subsection{Choice of pretraining strategy}
We provide an overview of the effect of pretraining choices in Figure \ref{fig:ptstrategy}, and analyze the choice of pretext task, effect of model size, and training duration, as well as differences between the two tracks.

\subsubsection{Pretext task} We group all pretext tasks into three overarching categories: \textit{local} objectives provide supervision on a voxel level, which for all methods in FOMO25 are variants of the MAE. \textit{Global} objectives provide supervision on a sequence or subject level and include variants of contrastive learning and supervised learning. \textit{Hybrid} objectives combine \textit{local} and \textit{global} objectives, which is commonly MAE + Contrastive. This also includes \textit{DINOv2}~\cite{Oquab2023}, which combines a global self-distillation loss on the \texttt{cls}-token with local self-distillation on the patch-tokens. 

To evaluate the pretext tasks invariant to absolute performance, Figure \ref{fig:ptstrategy} (g-i) considers the rank of each model across tasks. The large variation in rank across tasks for the same models shows that no single foundation model consistently performs well on one task. Figure \ref{fig:ptstrategy} (a-e) considers the difference between ranks over task types grouped by SSL objective type. We find that local objectives favor segmentation over classification, with no notable difference between either segmentation and regression or classification and regression. Hybrid objectives tend to favor classification and regression over segmentation, while there is no clear differentiation between classification and regression tasks. Global objectives surprisingly seem to favor segmentation over classification, and classification over regression, with no clear favorite between segmentation and regression. 

Figure \ref{fig:ptstrategy} (f) examines the effect of considering each sequence I.I.D. or trying to learn inter-sequence features during pretraining. We find that while the median rank is similar, the distribution of methods slightly favors methods that do not consider sequences I.I.D.

\subsubsection{Model size and training duration}
Figure \ref{fig:ptstrategy} (d) considers the overall rank compared to the size of their encoder. We find a slight upwards trend, indicating that larger models seem to benefit more from pretraining, however the correlation is not strong. Figure \ref{fig:ptstrategy} (e) considers the overall rank compared to the duration of pretraining, as measured by total number of samples during pretraining. Interestingly, we find that training for longer does \textit{not} yield meaningful improvements, indicating that methods saturate early. 

\subsection{Comparing the Open and Method Tracks}

Since the setup of the Open Track allowed participants to submit any foundation model, we provide a comparison in Figure \ref{fig:track_comparison}. We find that the Open Track, in general, used larger datasets than the Method Track (with \teamtagopen{tcRaidium}Curia (Raidium) using a dataset estimated at >1M scans, covering both CT and MRI), and larger models on more GPUs. Notably, all but two methods in the Open Track used a CNN, while this was only true for $50\%$ of models in the Method Track.

\subsection{Tuning}

\begin{figure}[!t]
    \centering
    \includegraphics[width=\linewidth]{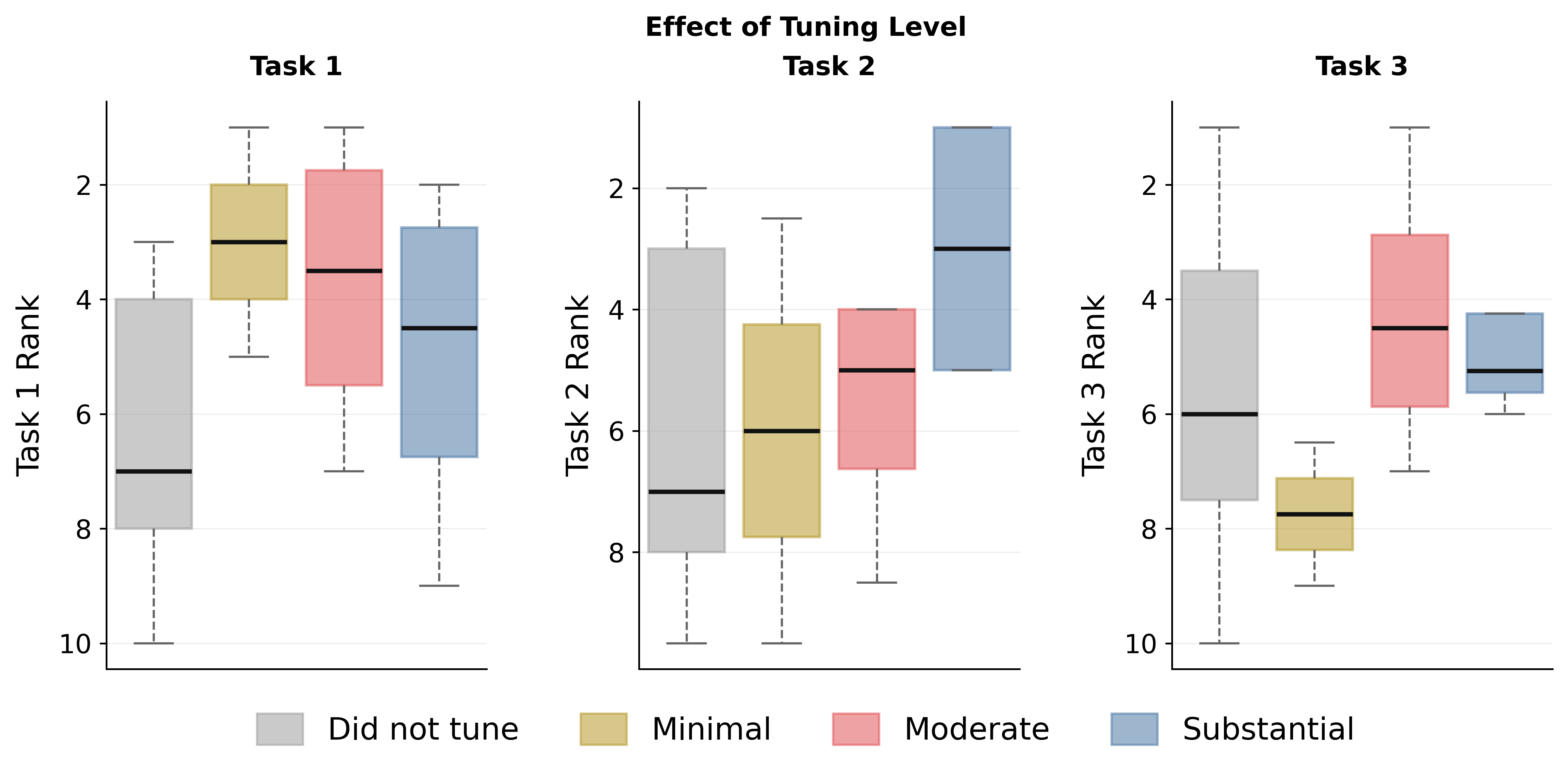}
    \caption{\textbf{Impact of hyperparameter tuning on final rank.} Categorization of the extent of hyperparameter tuning done for pretraining hyperparameters compared to the final rank on each task.}
    \label{fig:tuning}
\end{figure}

Participants reported their pretraining hyperparameter tuning strategies, which we classified as "Did not tune," "Minimal," "Moderate," or "Substantial." Figure \ref{fig:tuning} shows the effect of these strategies on task performance. On average, methods that did not tune performed worse. Minimal tuning was the best on classification tasks, indicating that these tasks are more susceptible to overfitting, especially in a few-shot, out-of-domain setting, due to the image-level signal. Substantial tuning benefited segmentation, but notably hurt regression tasks. No tuning level consistently benefited all tasks, suggesting the best tuning approach varies by task.

\subsection{Augmentations}

We provide an analysis of the effect of different augmentation strategies in Figure \ref{fig:augmentation}. We note that the total number of augmentations used seems to, on average, negatively impact classification and regression tasks, while the number of spatial augmentations used positively affected segmentation performance.

\section{Discussion}



\subsection{Pretraining boosts generalization}
The FOMO25 challenge was designed to provide a useful setting for examining the utility of foundation models on clinical data, with an emphasis on domain shift and limited annotations. Supervised baselines in our study exhibit poor few-shot and out-of-domain performance, reflecting their tendency to degrade when training and test distributions differ -- limitations which are inherent to their reliance on labels restricting the data distribution. Instead, the performance of foundation models can surpass not only supervised models trained on the same data, but also supervised models trained on in-domain data, showing that a well-tuned foundation model can bridge a cross-continent domain gap in a few-shot setting.

At the same time, supervised approaches remain highly competitive when large labeled datasets are available, as demonstrated by the \teamtagopen{tcFOMO2JOMO}FOMO2JOMO open-track brain age submission. Since the FOMO25 setup only benchmarks models on their few-shot performance, the results do not provide any indication on the utility when abundant labels are available. Taken together, while these results support the growing view that foundation models offer a credible path toward more robust clinical deployment, the utility of foundation models still depends on the problem and the need for out-of-domain or few-shot performance.

\subsection{Diversity and immaturity of current pretraining approaches}

The submissions to FOMO25 included a wide range of backbones, SSL objectives, and data preprocessing approaches, reflecting a field that remains exploratory. While such breadth is scientifically useful, it also indicates that there is not yet a mature consensus on how brain MRI foundation models should be pretrained and adapted. A core observation is that all models experienced a high degree of variation in their ranking across tasks. The comparisons of pretraining objectives suggest that the design choice of pretext task matters: local objectives (MAE) tended to perform better for segmentation tasks, whereas hybrid objectives generally yielded stronger performance on classification and regression. Augmentation choices further reinforced this pattern: spatial augmentations improved segmentation, while excessive augmentations sometimes reduced performance on classification and regression tasks.

These observations suggest that downstream performance currently depends on the alignment between the pretraining task and the target task. More specifically, the balance between local and global information in the pretraining objective appears to be an underexplored but consequential design parameter, and while many teams utilized hybrid objectives, the balance seems to overly favor classification and regression. This balance has been achieved in the natural imaging domain, where models such as DINOv3\cite{simeoni2025dinov3} exhibit competitive performance on both classification and segmentation tasks. Although these comparisons are observational, they provide useful guidance for future work in designing new pretraining strategies that better balance local and global information.

Contrary to findings in general computer vision \cite{He2022,woo2023convnext}, associations between model size, training duration, and downstream performance were weak. Larger models showed at most a modest advantage, and longer pretraining did not consistently translate into better results. This nuance is important: the field’s uncertainty about scaling is not because larger models or more compute have been shown to fail, but because it remains unclear how scale should interact with objective design, data composition, and adaptation strategy.

The submissions also reveal trends in architectural choices. Convolutional architectures remain dominant, and no submissions implemented 3D self-distillation methods \cite{caron2021dino,zhou2021ibot}, despite their growing use in large-scale visual representation learning. This likely reflects the longstanding convention of CNNs in medical imaging and the relative scarcity of established 3D implementations for newer training paradigms. Overall, these findings suggest that scaling, architecture, and advanced training paradigms remain underexplored in medical foundation models.

\subsection{Implications for future research directions}
Prior work has found only modest gains from self-supervised learning on large, research-grade benchmarks \cite{Wald2025b}. Our findings suggest that this may partly reflect benchmark design rather than the limited utility of pretraining itself. In brain image analysis, evaluation has largely centered on segmentation tasks, research-grade datasets with hundreds of labeled sessions, and in-domain testing. In such settings, supervised learning is a strong baseline, and downstream improvements from pretraining are naturally constrained, since models are effectively judged on how well they improve over already high-quality segmentation masks.

By introducing domain shift, clinical artifacts, and few-shot conditions, FOMO25 evaluates models on their ability to learn representations that generalize beyond narrowly curated datasets. The fact that no submitted method approached task saturation indicates that these benchmarks are informative and remain highly relevant for assessing pretraining quality in clinical settings.

Given the observed interaction between pretraining objectives and task performance, objectives should be selected with the downstream task family in mind. Similarly, scaling models without reconsidering objectives and data composition appears unlikely to yield substantial gains, as suggested by the weak associations between model size, training duration, and performance in our results. Finally, underexplored avenues such as 3D self-distillation and transformer-based architectures represent promising directions for future work. We expect future editions of the FOMO challenge to provide opportunities to test these insights on new tasks and datasets.

\section{Limitations}

\subsection{Data}

We note that some task-specific data characteristics may affect evaluation and should be considered when interpreting the results. For Task 2, meningioma characteristics may differ between the Indian finetuning cohort and the Danish validation and test cohorts, including typical location and size distributions, resulting in a clinically realistic but challenging transfer setting. In addition, segmentation labels were defined on FLAIR images, but complementary sequences were aligned to FLAIR using automated co-registration without manual correction. For clinical scans with a limited field of view, imaging artifacts, or low spatial resolution, co-registration quality may be reduced, potentially limiting the utility of non-FLAIR sequences for affected cases. For Task 3, finetuning cases were drawn from a cohort reported to have no major brain abnormalities. For validation and test datasets, cases were selected using ICD-10 codes to exclude major neurological or psychiatric disorders (cf.~\ref{app:Datasets}). These cohorts should not be interpreted as population-based healthy aging samples, but rather as clinically imaged individuals without reported brain pathology. While this design limits generalization to normative population brain aging, it evaluates representation transfer under realistic clinical sampling.

\subsection{Inference-only evaluation}
A key limitation of the challenge design is that evaluation was performed on participant-submitted inference pipelines rather than on a standardized finetuning protocol. This allowed teams to explore adaptation strategies such as LoRA~\cite{Hu2022} and other task-specific tuning recipes, which increases practical relevance, but it also confounds the interpretation of results because pretraining effects cannot be cleanly separated from finetuning choices. As a result, performance differences may reflect not only representation quality, but also differences in optimization, hyperparameter search, engineering effort, and computational resources. This design further reduces reproducibility, complicates attribution of negative results, and increases the risk of benchmark-specific overfitting.

\subsection{Track design}
A further limitation concerns the distinction between the Open and Method Tracks. In principle, this split was intended to assess the value of unrestricted external pretraining data relative to pretraining restricted to FOMO60K. In practice, however, the comparison was less informative than anticipated, as many Open-track submissions were based on models developed for a closely related challenge with a broadly similar pretraining scale. As a result, the two tracks may not have differed enough in effective data scale or diversity to reveal the full impact of unrestricted pretraining. More broadly, the track design conflates data access, data composition, and training methodology, making it difficult to attribute performance differences specifically to the use of external data. This issue may become more pronounced in future editions as pretraining datasets continue to grow in size and heterogeneity.

\subsection{Task choices}
Each downstream task category, including classification, segmentation, and regression, was represented by a single task. Apparent category-level effects are therefore inseparable from the specific properties of the selected tasks. For instance, a favorable result for one pretraining objective on segmentation may reflect the structure, label quality, or difficulty of that particular benchmark rather than a general advantage for segmentation tasks overall. The absence of multiple tasks per category limits within-category replication and weakens the generalizability of conclusions about how different pretraining strategies transfer across downstream task families.

\section{Conclusion}

FOMO25 was designed to evaluate whether foundation models for brain MRI can deliver practical value under the conditions that matter clinically: limited labels, heterogeneous acquisitions, and substantial domain shift. The results indicate that they can. Across tasks, self-supervised pretraining consistently improved adaptation to clinical data, and the strongest out-of-domain foundation models exceeded not only supervised baselines trained on the same pretraining source, but in some cases also specialist models trained in-domain.

At the same time, the results indicate there is no universally optimal recipe for foundation models in neuroimaging, and small models were competitive, with unclear benefits from scaling model size and training length. More broadly, FOMO25 highlights the need to benchmark foundation models on realistic clinical data rather than exclusively on curated research datasets. By centering evaluation on few-shot adaptation and out-of-domain transfer, the challenge provides a more stringent and clinically relevant picture of current capabilities. Overall, FOMO25 shows that self-supervised foundation models for brain MRI are no longer merely a promising research direction; they are becoming a credible alternative to task-specific supervised models in clinically realistic settings.

\section*{Acknowledgements}
The organisers were supported by the Pioneer Centre for AI (DNRF grant nr. P1); Lundbeck Foundation (grant number R449-2023-1512); Danish Data Science Academy (funded by the Novo Nordisk Foundation NNF21SA0069429 and Villum Fonden 40516); DIREC project EXPLAIN-ME (9142-00001B); Lundbeckfonden grant nummer R383-2022-285 


\teamtagmethod{tcAshash}ashash acknowledges support from Institute for Information \& communications Technology Promotion(IITP)
grant funded by the Korea government(MSIT) (No.00223446). 
\teamtagmethod{tcFOMO2JOMO}FOMO2JOMO acknowledges support from Research supported by the Swiss National Science Foundation (CRSII5\_202276/1) (Jaume Banus), the Hasler Foundation Responsible AI program (MSxplain) and the Research Commission of the Faculty of Biology and Medicine (CRFBM) of UNIL (Nataliia Molchanova), the NVIDIA project “Deep Generative Models to Learn Causal Mechanisms for Automated Multiple Sclerosis Assessment with Brain MRI” through the Academic Hardware Grant Program (Pedro M. Gordaliza), the Walloon region under grant No. 2010235 (ARIAC by digitalwallonia4.ai) (Maxence Wynen). The team acknowledges access to the facilities and expertise of the CIBM Center for Biomedical Imaging, a Swiss research center of excellence founded and supported by CHUV, UNIL, EPFL, UNIGE, and HUG. Training our models also benefited from computational resources made available on Lucia, the Tier-1 supercomputer of the Walloon Region, infrastructure funded by the Walloon Region under the grant agreement no 1910247.
\teamtagmethod{tcMcGill}McGill\_PVG was supported by the Natural Sciences and Engineering Research Council of Canada, Fonds de Recherche du Quebec: Nature et Technologies, the Canadian Institute for Advanced Research (CIFAR) Artificial Intelligence Chairs program, Calcul Quebec, the Digital Research Alliance of Canada, the Vadasz Scholar McGill Engineering Doctoral Award, Mila - Quebec AI Institute, the International Progressive MS Alliance and the MS Society of Canada.
\teamtagmethod{tcMIPLAB}MIPLAB was supported by the Canada Research Chairs Program, Canadian Neuroanalytics scholar (CNS) program, and VAST Post-doctoral Fellowship.
\teamtagopen{tcLatent}LatentCampus acknowledges the funding from the project ROBUST: Trustworthy AI-based Systems for Sustainable Growth with project number KICH3.LTP.20.006, which is (partly) financed by the Dutch Research Council (NWO), Siemens Healthineers, and the Dutch Ministry of Economic Affairs and Climate Policy (EZK) under the program LTP KIC 2020-2023.
\teamtagmethod{tcTUKE}TUKE was funded by the EU NextGenerationEU through the Recovery and Resilience Plan for Slovakia under the project No. 09I03-03-V04-00394
\teamtagopen{tcCAMI}CAMI\_SIAT was supported by the National Key R\&D Program of China (Grant No. 2024YFF0507502)  

\section*{Author contributions}

Challenge conception and design: A.M., S.C., M.N. Challenge coordination and project management: A.M., S.C., S.N.L., and M.N. Pretraining dataset curation and release: A.M., S.C., and J.M. Finetuning, validation, and test data curation: A.M., S.C., V.N., C.H.K., E.J.S., and A.P. Manual annotation of Task 1 and Task 2 cases: V.N., C.H.K. Evaluation infrastructure and secure cluster setup: A.M., S.C., P.R.G., and S.N.L. Container evaluation and submission handling: A.M., S.C., P.R.G., and S.N.L. Baseline model implementation and training: A.M., J.A., P.R.G., and S.N.L. Statistical analysis and figures: A.M., S.C., and M.N. Manuscript drafting: A.M. and S.C. Supervision and funding acquisition: P.L., M.M.G., M.B., M.E.B., J.E.I., and M.N. Participating-team contributions: each participating team designed, implemented, pretrained, finetuned, and submitted their own method as described in~\ref{sec:submissions_overview}; team members from each submitted method are listed as co-authors in alphabetical order. All authors reviewed and approved the final manuscript.

\section*{Ethical approvals}
Validation and test datasets were approved by the Danish Patient Safety (Styrelsen for Patientsikkerhed, approval \#31-1521-257) and the Danish Data Protection Agency (Datatilsynet, approval \#P-2020-320).

\section*{Declaration of generative AI and AI-assisted technologies in the manuscript preparation process}

During the preparation of this work, the authors used Claude (Anthropic) and ChatGPT (OpenAI) for light language editing. After using these tools, the authors reviewed and edited the content as needed and take full responsibility for the content of the published article.

\bibliographystyle{elsarticle-num-names}
\bibliography{references} 

\appendix

\section{Dataset Details}
\label{app:Datasets}

This appendix details subject demographics, MRI sequences, preprocessing, and labeling protocols for all FOMO25 finetune, validation, and test datasets. Table~\ref{tab:dataset-summary} summarizes the key characteristics of each dataset split.

\begin{table*}[t]
\centering
\footnotesize
\setlength{\tabcolsep}{6pt}
\begin{tabularx}{\textwidth}{l l c c c l l c}
\toprule
\textbf{Task} & \textbf{Split} & \makecell{\textbf{\# Cases}} & \makecell{\textbf{Data}\\\textbf{Origin}} & \makecell{\textbf{Age}\\\textbf{(mean $\pm$ SD)}} & \makecell{\textbf{Sex}\\\textbf{(M/F)}} & \textbf{Sequences} & \makecell{\textbf{Slice thickness}\\\textbf{(mean $\pm$ SD)}} \\
\midrule
\multirow{3}{*}{Task 1 - Infarct Classification}
 & Finetuning & 21  & India & 61.52 $\pm$ 16.27 & 16 / 5 & \begin{tabular}[t]{@{}l@{}}FLAIR, DWI (ADC+b1000), \\ T2*(5) / SWI(16)\end{tabular} & 4.50 $\pm$ 0.89\\
 & Validation & 80  & Denmark & 63.17 $\pm$ 17.27 & 39 / 41 & \begin{tabular}[t]{@{}l@{}} FLAIR, DWI (ADC+b1000), \\ T2*(15) / SWI(65)\end{tabular} & 4.57 $\pm$ 1.73\\
 & Test       & 320 & Denmark & 64.30 $\pm$ 17.37 & 161 / 159 & \begin{tabular}[t]{@{}l@{}}FLAIR, DWI (ADC+b1000), \\ T2*(59) / SWI(261)\end{tabular} & 4.57 $\pm$ 1.77\\
\midrule
\multirow{3}{*}{Task 2 - Meningioma Segmentation} 
 & Finetuning & 23  & India & 48.27 $\pm$ 22.04 & 8 / 15 & \begin{tabular}[t]{@{}l@{}}FLAIR, DWI (b1000), \\ T2*(15) / SWI(8)\end{tabular} & 4.61 $\pm$ 0.78 \\
 & Validation & 25  & Denmark & 65.36 $\pm$ 13.54 & 7 / 18 & \begin{tabular}[t]{@{}l@{}}FLAIR, DWI (b1000), \\ T2*(2) / SWI(23)\end{tabular} & 4.11 $\pm$ 2.04\\
 & Test       & 107 & Denmark & 67.01 $\pm$ 13.19 & 30 / 77 & \begin{tabular}[t]{@{}l@{}}FLAIR, DWI (b1000), \\ T2*(16) / SWI(91)\end{tabular} & 4.28 $\pm$ 2.03\\
\midrule
\multirow{3}{*}{Task 3 - Brain Age Regression} 
 & Finetuning & 200 & USA & 61.85 $\pm$ 15.09 & 105 / 95 & T1w, T2w & 5.07 $\pm$ 1.84\\
 & Validation & 200 & Denmark & 46.54 $\pm$ 16.23 & 119 / 81 & T1w, T2w & 3.88 $\pm$ 2.02\\
 & Test       & 800 & Denmark & 48.72 $\pm$ 17.03 & 473 / 327 & T1w, T2w & 3.88 $\pm$ 1.90\\
\bottomrule
\end{tabularx}
\caption{Summary of datasets used in FOMO25. For each task, we report the split, number of cases, data origin, age, sex, sequences included, and slice thickness. Numbers in parentheses indicate cases per sequence type for T2*/SWI. Slice thickness reported for original data before co-registration, averaged across all sequences.}
\label{tab:dataset-summary}
\end{table*}

\subsection{Finetuning, Validation and test datasets acquisition and preprocessing}

Scans were acquired on a heterogeneous mix of clinical MRI scanners spanning both 1.5T and 3T systems and multiple vendors; specific scanner models were not standardized across cases. 
All scans were first reoriented to RAS orientation and visually checked for preprocessing errors, including incorrect orientation, corrupted files, or empty images. For Task 1 and Task 3, scans were skull-stripped using SynthSeg~\cite{Billot2023} and affinely co-registered within each case to the highest-resolution scan using FreeSurfer v7.4.1 (\texttt{mri\_coreg})~\cite{Fischl2012}. For Task 2, scans were defaced with FreeSurfer v7.4.1 (\texttt{mideface}) to preserve extracerebral structures near the meningioma, and all sequences were affinely co-registered to the FLAIR scan, which served as the primary reference for manual annotation. All preprocessing steps were applied consistently across the finetuning, validation, and test datasets.

\subsection{Labeling protocol}

\noindent
\textbf{Task 1}: Finetuning cases were collected by Cerebriu and were reported to either contain an infarct or be free of infarcts. For validation and test datasets, positive cases were selected based on ICD-10 code I63 and negative cases based on the absence of I63 and I64 codes.  All subjects were $\ge$18 years at the time of the MRI scan. All cases were visually inspected by two expert annotators, V.N. (MD, PhD in neurology, $\ge$1 year of neurology residency) and C.H.K. (MD, PhD in radiology, $\ge$1 year of radiology residency), who consulted radiology reports in cases of uncertainty. DWI (b1000) and ADC images were used as the primary reference, with additional sequences consulted when needed. For the finetuning set, cases with disagreements between annotators were discussed until a consensus was reached. For validation and test datasets, if consensus was not reached between the annotators, the case was excluded. During annotation, images were also checked for incorrect sequences and preprocessing errors.

\noindent
\textbf{Task 2}: Finetuning cases were collected by Cerebriu and reported to contain meningiomas. For validation and test datasets, positive cases were initially retrieved based on ICD-10 codes D32.0-D32.9, excluding D32.1, and subjects were $\ge$18 years at the time of the MRI scan. All retained cases were manually annotated by the same expert annotators as Task 1 (V.N. and C.H.K.) following a protocol similar to the BraTS Meningioma Segmentation Challenge~\cite{Labella2024}. Annotations included meningioma tails but excluded peritumoral edema, with FLAIR scans used as the primary reference and additional sequences consulted when needed. Unlike BraTS, binary masks were generated instead of multi-label masks. Cases in which the meningioma could not be reliably delineated using the available sequences were excluded. For the Cerebriu finetuning cases, existing segmentations were available and were reviewed and refined by the annotators. V.N. labeled 12 cases for finetuning, 12 for validation, and 57 for testing, while C.H.K. labeled 11, 13, and 50 cases, respectively. During annotation, scans were also checked for incorrect sequences and preprocessing errors.
Inter- and intra-annotator variability were not quantified. Likely residual errors stem from meningioma tail delineation and partial-volume effects on low-quality scans.

\noindent
\textbf{Task 3}:  Finetuning cases were selected from a subset of the dataset in~\cite{Iglesias2023} acquired in Boston; these subjects were reported to have no major brain abnormalities, and no ICD-10 codes were used for selection. For validation and test datasets, subjects were selected following the protocol in~\cite{Cerri2026b} (Population B), using ICD-10 codes to exclude individuals with psychiatric diagnoses (F00-F99, G30), neurological comorbidities (i.e., disorders of the nervous system G00-G99, except G560 and G562; cerebrovascular diagnoses I60-I69, E236E; CNS infections A022C, A066, A17, A229C, A321, A390, A504, A514B, A521A-B, A548A, A80-A89, B003-B004, B010-B011, B020-B021, B050-B051, B060, B261-B262, B375, B451, B582, B602, E236A; traumatic brain injury S060-S069; skull fractures S020, S021, S027, S029; malignant cancers C00-C99; benign neoplasms of the nervous system D32-D33; HIV B20-B24), and those with brain-related medications (N03-N07, except N05AD08 and N07BA01). All subjects were $\ge$18 years at the time of the MRI scan. S.C. visually inspected all scans to check for incorrect sequences or gross brain abnormalities. Labels corresponded to the subject's chronological age at scan time, in years (floored).

\begin{figure*}[!ht]
    \centering
    \includegraphics[width=\linewidth]{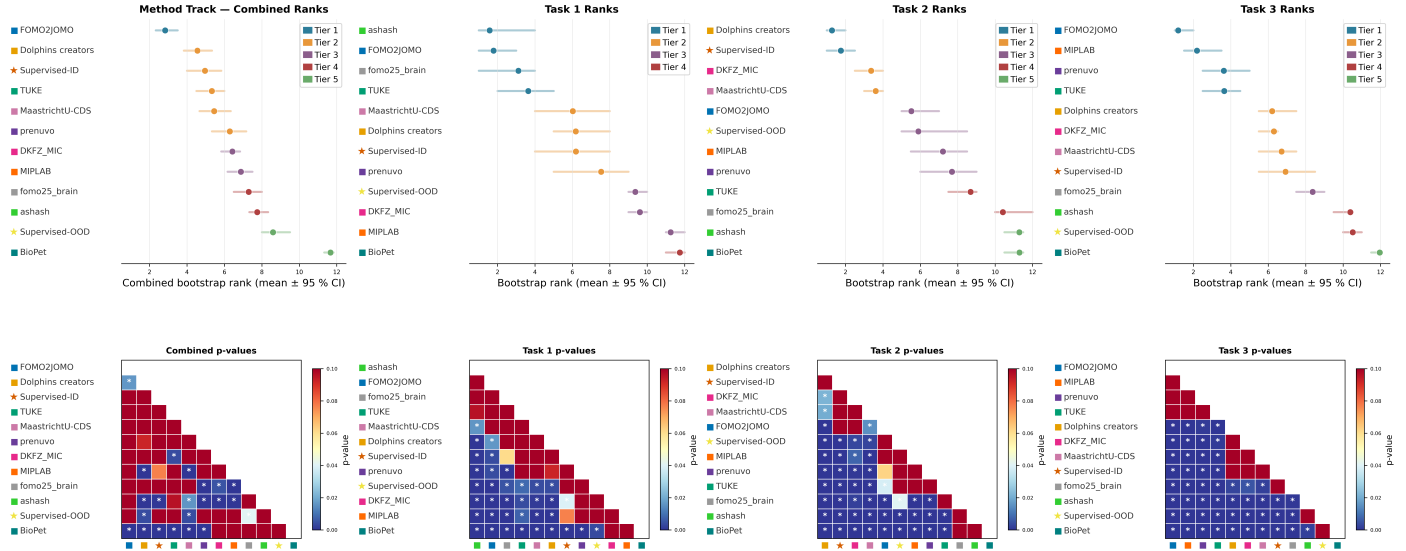}
    \caption{Statistical analysis for the Method Track. Top row: bootstrap rank distributions for the combined ranking and each individual task, with independently computed performance tiers. Bottom row: pairwise permutation test p-value matrices (lower triangular); asterisks denote statistically significant differences (p < 0.05). Teams are ordered by mean rank within each task.}
    \label{fig:method_stats_analysis}
\end{figure*}

\begin{figure*}[!ht]
    \centering
    \includegraphics[width=\linewidth]{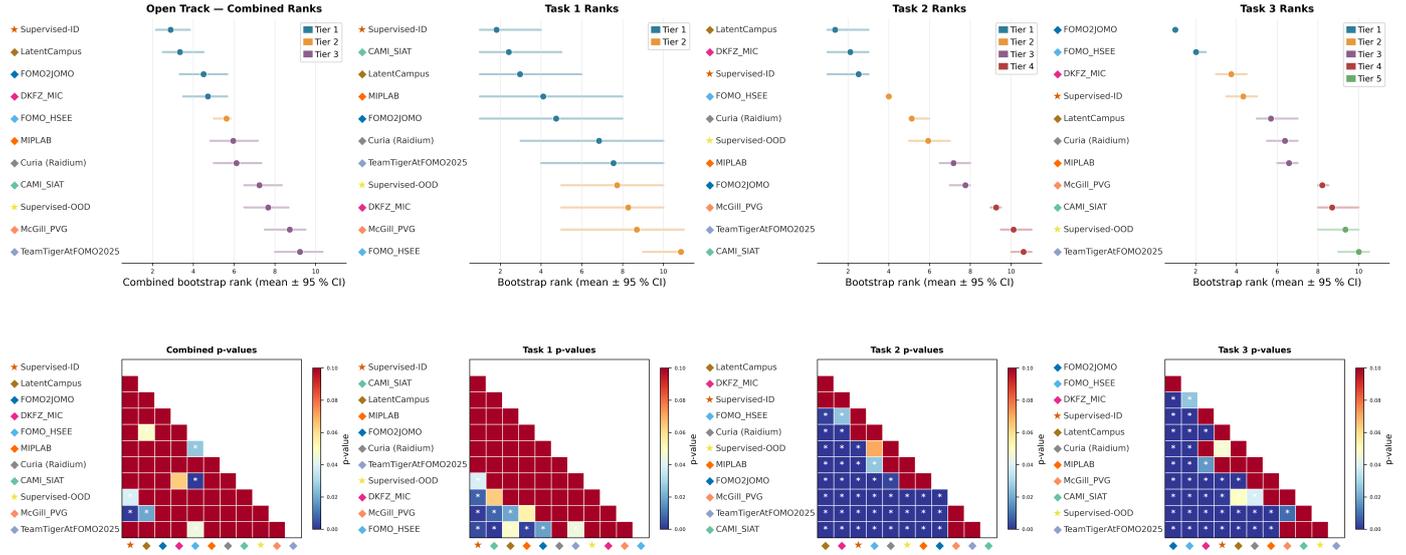}
    \caption{Statistical analysis for the Open Track. Top row: bootstrap rank distributions for the combined ranking and each individual task, with independently computed performance tiers. Bottom row: pairwise permutation test p-value matrices (lower triangular); asterisks denote statistically significant differences (p < 0.05). Teams are ordered by mean rank within each task.
    }
    \label{fig:open_stats_analysis}
\end{figure*}

\section{Participant Method Summaries}
\begin{figure*}[!t]
    \centering
    \includegraphics[width=\linewidth]{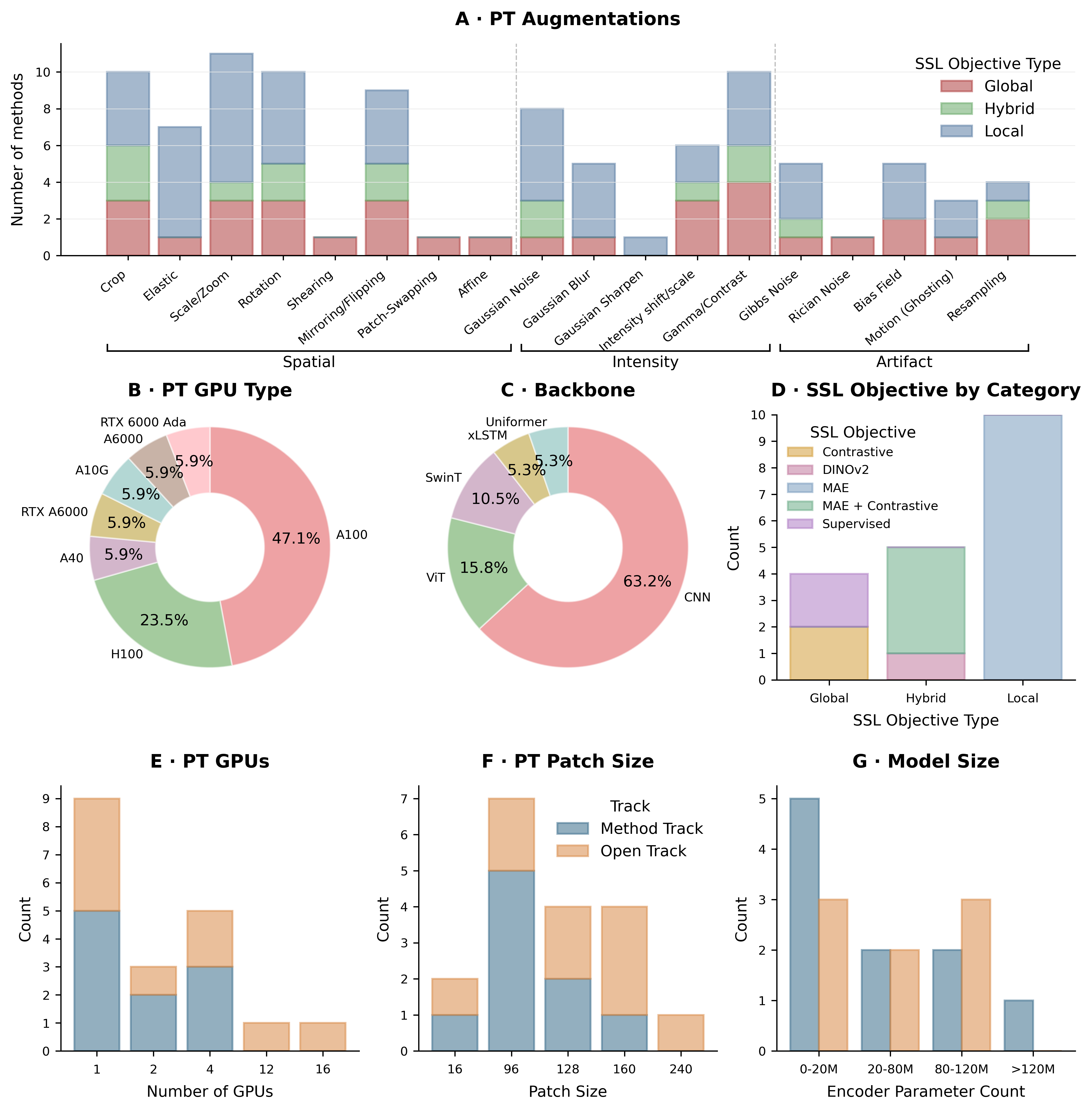}
    \caption{\textbf{Overview of submitted methods}. (A) Pretraining augmentations used across all submissions, grouped by spatial, intensity, and artifact categories and colored by SSL objective type (Global, Hybrid, Local). (B) Distribution of GPU types used for pretraining. (C) Backbone architectures, showing the proportion of CNN, Swin Transformer, Uniformer, Vision Transformer, and xLSTM models. (D) Count of SSL objective types by category. (E) Number of GPUs used during pretraining. (F) Pretraining patch sizes, stratified by track. (G) Encoder parameter counts. Panels E–G separate submissions per track.}
    \label{fig:state-of-the-union}
\end{figure*}

This section describes all eligible submissions received for the challenge. We provide an overview of high-level characteristics in Figure \ref{fig:state-of-the-union}.

\label{sec:submissions_overview}

\subsection{Method Track}
\subsubsection{\teamtagmethod{tcFOMO2JOMO}FOMO2JOMO}
\textbf{Pretraining.} 3D CNN multi-modal UNet-style VAE~\cite{Kingma2013} with 18.9M parameters with SmoothL1 masked reconstruction both on a scan and a subject level (sampling pairs of scans per subject) using a masking ratio of 0.2, a cosine-distance contrastive term on subject-level representations, and KL regularization on both subject-level and modality-specific latents \cite{banus2026} (with a 10-epoch KL warm-up). Training used patches of size $64\times64\times64$ with batch size $4$ using AdamW~\cite{Loshchilov2017} (LR $10^{-4}$, weight decay $0.01$) under a cosine annealing schedule with 5 warm-up epochs for 100 epochs on a single NVIDIA A100 GPU.

\textbf{Finetuning.} All tasks were trained for 150 epochs per fold, with no augmentations and a five-fold ensemble. 
\textbf{Classification.} Decoder was a two-layer MLP with SiLU activation function, which was fed the concatenated modality latent projections plus the averaged subject-level latent, optimizing cross-entropy with class weights $[0.4,0.6]$ and averaging ensemble probabilities (with an additional training-time rule that patch labels are set to $0$ when a positive-label patch contains no lesion). \textbf{Segmentation.} UNet-style decoder producing per-modality predictions fused by a learnable fusion head with encoder skip connections and the last bottleneck replaced by the latent projection, optimizing a weighted cross-entropy and Tversky loss with ensemble probability averaging and post-processing by retaining only the largest connected component. \textbf{Regression.} Decoder was an MLP optimized with SmoothL1 with model selection by validation Mean AE and an ensemble that takes the median predicted age.

\subsubsection{\teamtagmethod{tcDolphins}Dolphin\_creators}
\textbf{Pretraining.} A $\sim$45M-parameter Bi-Directional xLSTM-UNet (3D UNet backbone with bidirectional xLSTM~\cite{Beck2024} modules in the encoder and bottleneck) using a masked autoencoder objective with mean squared error (MSE) reconstruction loss. Each sequence was treated i.i.d. applying random rotations/flips/scaling with Gaussian blur and additive Gaussian noise, using $96\times96\times96$ patches with batch size 2, and optimizing with AdamW under cosine annealing from $2e^{-4}$ for $200$ epochs on a NVIDIA RTX A6000 (48 GB), taking $\sim$1 week and selecting the checkpoint with the lowest validation reconstruction loss.

\textbf{Finetuning.}
\textbf{Classification.} Global Active Pooling and then MLP head with dropout trained with focal loss using Adam~\cite{Kingma2014} and reduce-on-plateau-lr from $1e^{-4}$, with standard augmentations, early stopping with best checkpoint by validation accuracy. 5-fold cross-validation ensembling and 3-axis flip test-time augmentation (8 flips) via probability averaging. \textbf{Segmentation.} UNet-style decoder and Dice+cross-entropy loss, AdamW with cosine annealing from $2e^{-4}$ for 1000 epochs with best checkpoint by validation Dice. Training used standard augmentations. Inference uses a sliding window with Gaussian weighting/mirroring and connected-component filtering, and 5-fold ensemble inference (with deep supervision tested but disabled in the final model). \textbf{Regression.} Adaptive pooling and then MLP head trained using MSE loss with AdamW and cosine annealing from 1e-4 for up to 500 epochs with early stopping and best checkpoint by lowest validation Mean AE, using fold-wise ensemble averaging.

\subsubsection{\teamtagmethod{tcTUKE}TUKE}
\textbf{Pretraining.} SwinUNETR~\cite{Hatamizadeh2021} with 18.3M parameters (and lightweight encoder at 4M parameters) optimized by a weighted sum of a masked reconstruction objective (mask ratio 0.6) and a MoCo v2-style multimodal contrastive objective~\cite{chen2020a} (InfoNCE with a momentum-updated key encoder and a negative-sample queue). Sequences from the same subject were treated as independent single-channel volumes, with positive pairs formed across modalities. Augmentations were a combination of spatially correlated geometric augmentations and intensity and noise augmentations. Patch size $96\times 96\times 96$ per GPU, batch size 10 with 5-step gradient accumulation. Optimized with AdamW (LR $10^{-4}$ with cosine annealing) using 4$\times$ NVIDIA A100 GPUs, yielding a global batch size of 200, for approx 4 days. The checkpoint was selected from the last epoch.

\textbf{Finetuning.} For all tasks, the backbone was frozen, and only LoRA~\cite{Hu2022} adapters were trained, and all standard augmentations were applied.
\textbf{Classification.} Multi-scale pooled classification head with late fusion across sequences, AdamW with cosine annealing with warm restarts, early stopping on validation AUROC, best-AUROC checkpoint selection, and K-fold ensemble plus test-time augmentation majority voting. \textbf{Segmentation.} Late fusion by averaging per-sequence logits. Simple post-processing via the largest connected component. \textbf{Regression.} Early-fuse sequences as input channels. Finetune the full encoder. Use a pooling+MLP regression head. Patch-based training with age-balanced sampling. Loss targets Mean AE and age-dependent bias. Early stopping on bias-corrected Mean AE. Applied linear bias correction after training.

\subsubsection{\teamtagmethod{tcMaastricht}MaastrichtU-CDS}
\textbf{Pretraining.} SwinUNETRv2~\cite{He2023} with 22M Parameters (20M in Encoder) using MoCo v2 objective plus an auxiliary 5-class modality classification objective trained with weighted cross-entropy. Contrastive pairs were formed on a session level. Extensive spatial (flips/rotation/scaling/shearing/cropping) and intensity (scaling/Rician noise/Gaussian smoothing/bias field/Gibbs noise/contrast/resampling artifact emulation) augmentations are applied independently to query and key. Data curation via skull-stripping (filtered scans with less than 12cm brain) and filtering (removed \texttt{scan\_*} sequences), yielding $41,900$ scans from $11,976$ sessions across $9,606$ individuals. Training used $128^3$  patches, batch size $4$, AdamW (LR $10^{-4}$) with cosine annealing and $5\%$ warmup over $100$ epochs on one H100 80GB GPU for $43.1$ hours, selecting the lowest-validation-loss checkpoint. 

\textbf{Finetuning.} Across all finetuning tasks, late-fusion is applied. 
\textbf{Classification.} Global Average Pooling (GAP) across modality-embeddings followed by an MLP. Deepest encoder level unfrozen after $70\%$ steps. Batch size $1$ with accumulation $4$ for 285,100 steps, stochastic weight averaging over the last $20\%$ of steps, and test-time flipping plus mean logit ensembling over $5$ seeds. \textbf{Segmentation.} fuses per-level features via a learnable $(n_{\text{levels}} \times n_{\text{modalities}})$ matrix into a lightweight Feature Pyramid Network-style decoder trained with Dice Cross-Entropy under an AdamW warmup+cosine annealing scheme. Deepest encoder-level unfreezing after $70\%$ for 540,000 steps, with sigmoid activation and $0.5$ thresholding, and the same $5$-seed flip-TTA mean logit ensembling. Data was skull-stripped and registered to MNI152 space. \textbf{Regression.} GAP over encoder features and then GAP across $8$ overlapping patches (2 per spatial dimension) before a learnable weighted modality average and an MLP trained with SmoothL1 loss using a staged unfreezing schedule across encoder layers for 240,000 steps with batch size $2$ and accumulation $4$, SWA over the last $20\%$ of steps, and mean ensembling of predicted age over $5$ seeds.

\subsubsection{\teamtagmethod{tcDKFZ}DKFZ\_MIC}
\textbf{Pretraining.} ResEnc-L CNN with 102M parameters pretrained on FOMO60k using a Masked Autoencoder (MAE) objective with MSE loss. Sequences sampled i.i.d. nnUNet spatial augmentations. No hyperparameter tuning was performed beyond the default \texttt{nnssl} setup~\cite{Wald2025b}. Training used $160^3$ patches, batch size 8, SGD with polynomial decay over 1000 epochs / 250 steps per epoch on four A100 40GB GPUs for 46 hours.

\textbf{Finetuning.} When the number of input channels at fine-tuning exceeds that at pretraining, weights from the first encoder layer are repeated accordingly.
\textbf{Classification.} Classification head is linear layer on encoder features. Training used $160^3$ patches, batch size 1 with 48-step gradient accumulation, AdamW with cosine annealing for 400 epochs / 5 steps per epoch on one A100 40GB. Extensive spatial and intensity augmentations are applied alongside 0.2 label smoothing. Five-fold cross-validation with best-validation-loss checkpoint selection; test-time mirroring (8 iterations) with class-probability ensembling over the top-3 cross-validation models.
\textbf{Segmentation.} Standard U-Net decoder trained with Dice + Cross-Entropy loss. Training used $160^3$ patches, batch size 2, SGD with polynomial decay for 150 epochs on one A100 40GB, using default nnUNet augmentations and postprocessing with five-fold cross-validation checkpoint ensemble.
\textbf{Regression.} Head is a linear layer, combining pooled features from multiple scales. Trained with MSE loss using T1 scans only. Training used $160^3$ patches, batch size 4 with 48-step gradient accumulation, SGD with cosine annealing for 1000 epochs on one A100 40GB. Regularization includes 0.5 dropout in residual connections, mixup, label smoothing, shakedrop, undecay, zero-initialized residuals, and squeeze excitation, with deep supervision. Test-time mirroring (8 iterations) with age-prediction averaging over a five-model cross-validation ensemble.

\subsubsection{\teamtagmethod{tcMIPLAB}MIPLAB}
\textbf{Pretraining.} The organizer-provided checkpoint (\texttt{unet\_b\_100}) was used, a 3D CNN U-Net with 20M parameters trained on FOMO60k via MAE with MSE loss.

\textbf{Finetuning.} All tasks used patch size $96^3$, batch size $2$, AdamW, base LR $10^{-4}$) with cosine annealing, layer-wise learning rates per encoder depth group (decay factor $2.5$), and extensive \texttt{batchgenerators}~\cite{Isensee2021} augmentation (elastic deformation, rotation/scaling, flipping, Gaussian noise/blur, brightness, gamma) applied simultaneously across all input modalities. 
\textbf{Classification.} Used organizer's \texttt{ClsRegHead}; cross-entropy loss; 21 epochs / 231 steps; $\sim$20 minutes per sweep; early stopping on training loss. Bayesian HP search over LR, layer-wise learning rate decay, and loss weighting. Trained on one RTX~3090 (24GB).
\textbf{Segmentation.} Standard U-Net Decoder. DiceCE loss, trained for 31 epochs with early stopping on validation loss, progressive encoder unfreezing from deepest blocks (at epoch 5) to input convolution (epoch 25); separate encoder/decoder LRs ($3\times10^{-5}$ / $3\times10^{-4}$) under a shared cosine schedule with newly unfrozen layers inheriting the current encoder LR; foreground oversampling probability 0.85; predictions restricted to within-brain-mask voxels. Trained on one V100 (16GB).
\textbf{Regression.} Organizer's \texttt{RegHead} extended with additional linear layers, batch normalization, and dropout; modified MSE loss with variance penalty to prevent mean-age collapse; age range split into 3 equally resampled bins; gradual unfreezing/refreezing strategy over $\sim$600 epochs for $\sim$2.5 hours; manual stopping; linear regression post-processing for age-level bias correction. Trained on one RTX~3090 (24GB).

\subsubsection{\teamtagmethod{tcPrenuvo}prenuvo}
\textbf{Pretraining.} A 3D Vision Transformer backbone (109M parameters, linear project decoder $\sim$4.2M) trained with a SimMIM-style masked image reconstruction objective~\cite{Xie2022} optimized with an L1 reconstruction loss. Each scan is treated i.i.d. using $96\times96\times16$-voxel volumes with $16\times16\times16$ patches and a 50\% masking ratio, no pretraining data augmentation, and training with batch size 32 under AdamW (learning rate 1e-5) and a WarmupCosineScheduler (1000 warmup steps) for 45 epochs on a single NVIDIA A10G GPU with mixed-precision (FP16).

\textbf{Finetuning.}
\textbf{Classification.} Mean-pooled patch embedding followed by a linear projection classifier, trained with a cosine-annealing-with-warm-restarts schedule, batch size 4 (validation 1), patch size $64\times64\times16$ with 0.25 overlap, progressive unfreezing at [50, 200, 500] epochs, class weights [0.1, 0.9], and checkpoint selection by maximum validation F1, with MSE loss and staged augmentations escalating from light flips/rotations to added Gaussian noise and contrast adjustment \textbf{Segmentation.} UNETR-style decoder (a 4-layer autoencoder with skip connections from transformer blocks 2/4/6/8) and L1 loss; \textbf{Regression.} mean-pooled linear projection head with the same training configuration as Task 1 and an MSE loss.

\subsubsection{\teamtagmethod{tcBrain}fomo25\_brain}
\textbf{Pretraining.} ViT (500M parameters, 450M in encoder), using MAE and cross-modality translation. Data curation to filter only to subjects with paired, registered multi-sequence scans (resulting in $8,477$ subjects, $10,580$ sessions, $31,556$ scans). Multiple sequences were handled by randomly forming modality pairs. Trained for $200$ epochs on 2$\times$ NVIDIA A100 for about 2 days, checkpoint selection based on best validation loss.

\textbf{Finetuning.}
\textbf{Classification.} linear head, patch size 16 with patch size 96, batch size 16, AdamW for 200 epochs using cross-entropy loss, using provided voxel-level labels to generate new labels (label 1 if a cropped patch contains $>20$ foreground voxels), sliding-window inference where the final probability is the second-highest patch prediction. \textbf{Segmentation.} Reusing original pretraining decoder, Dice loss plus weighted cross-entropy, otherwise the same optimizer and hyperparameters, with inputs resampled to isotropic 1mm. \textbf{Regression.} linear head, MSE loss, otherwise the same optimizer and hyperparameters, with resampling to isotropic 2 mm, padding/cropping to size 96, and patchwise predictions combined after inference.

\subsubsection{\teamtagmethod{tcAshash}ashash}
\textbf{Pretraining.} 3D CNN U-Net autoencoder with 6M parameters (encoder 3.5M parameters), using a multi-ratio MAE objective, where masking patch sizes are randomized in ranges $8$-$16$ and $32$-$48$. Sequences treated i.i.d, with all inputs interpolated to $128\times128\times128$ patches. Augmentations (Spatial flip, Gaussian noise, bias field, Gibbs noise, gamma transform, Gaussian smoothing, sharpening) were applied \emph{before} masking with the reconstruction target being the original images. Optimization used AdamW (initial LR $10^{-3}$, weight decay $0.05$) with cosine annealing and warm restarts, trained for $100$ epochs (stage 1, all images 1mm isotropic) plus $15$ epochs (stage 2, all images in $0.7\times0.7\times1$ mm spacing) at effective batch size $64$ (per-GPU batch $16$ across $4$ GPUs). Training used $4\times$ NVIDIA RTX 6000 Ada with bf16 mixed precision, for approximately a week (Stage 1) plus $\sim1$ day (Stage 2), saving checkpoints every 5 epochs without early stopping.

\textbf{Finetuning.} For all tasks, data was resampled to 0.7$\times$0.7$\times$1,mm spacing.
\textbf{Classification.} Multi-modal linear mixers fuse per-stage features followed by GeM pooling, layer normalization, and a linear classification head. BatchNorm statistics frozen. Trained with cross-entropy using AdamW (LR $5\times10^{-4}$, weight decay $0.01$) and cosine warmup restarts for 340 steps on $224\times288\times160$ patches, with spatial-flip augmentation and a 5-fold soft-voting ensemble. \textbf{Segmentation.} Pretrained U-Net decoder fusing modality pairs only at stages 5 and 4 with Conv-GroupNorm-ReLU mixers, using FLAIR-based skip connections at stages 1-3, and adding the stage-5 FLAIR features back to the fused representation via a residual connection to preserve FLAIR information. Trained with DiceFocalLoss using AdamW (LR $10^{-4}$, weight decay $0.01$) and cosine warmup restarts for 4000 steps on $192\times192\times128$ patches, with spatial flips and a 3-seed ensemble trained excluding a misaligned subject. \textbf{Regression.} mirrors classification task but replaces the head with a linear regression head and uses a Mean AE loss with a normalized scale divided by 100. Trained with AdamW (LR $5\times10^{-4}$, weight decay $0.01$) and cosine warmup restarts for 3200 steps on $256\times320\times192$ patches, with spatial flips, prediction clipping to $[18,100]$, and a 5-fold soft-voting ensemble stratified by age bins.

\subsubsection{\teamtagmethod{tcBiopet}biopet}
\textbf{Pretraining.} Used the organizer-provided pretrained U-net B model (22M parameters) trained with a MAE objective with masking probability $0.6$ and optimizes an MSE reconstruction loss. Each scan was treated i.i.d. during pretraining with an extensive augmentation scheme over $100$ epochs with $5$ warmup epochs.

\textbf{Finetuning.} For all finetuning tasks, the encoder was frozen using a single GTX 1080Ti.
\textbf{Classification.} Used patch size $160$ for $50$ epochs. \textbf{Segmentation.} trained with patch size $160$ for $50$ epochs using an asymmetric unified focal loss and mask augmentation via random dilation (Gaussian blur) \textbf{Regression.} trained with patch size $160$ for $100$ epochs.

\subsection{Open Track}
\subsubsection{\teamtagopen{tcLatent}LatentCampus}

\textbf{Pretraining.} 3D CNN pretraining setup based on a ResEnc-L encoder with ($\sim$102.35M parameters) trained with their Consistent View Alignment (CVA) objective~\cite{Vaish2025} that combines MAE with a Huber reconstruction loss, an ROI-based alignment term using a symmetrized cosine regression loss, and an optional symmetrized global contrastive NT-Xent term, aggregated as $$\mathcal{L} = \lambda_{\text{rec}} \, \mathcal{L}_{\text{rec}} \;+\; \lambda_{\text{consis}} \,
\mathcal{L}^{\text{sym}}_{\text{consis}} \;+\; \lambda_{\text{con}} \,
\mathcal{L}^{\text{sym}}_{\text{con}}.$$ Training data is the OpenMind dataset~\cite{Wald2025b}. Each sequence is treated i.i.d., uses random 3D cropping with $40$-$80\%$ overlap (for CVA), rotations, scaling, and flips plus intensity shifts, additive Gaussian noise, and gamma adjustments, with patch size $128^3$, masking ratio $70\%$ over $250$ epochs at $250$ steps/epoch on a single NVIDIA A40 GPU for approximately $2$ days, selecting the last checkpoint for downstream use.

\textbf{Finetuning.} Stem and decoder are initialized from scratch.
\textbf{Classification.} Single linear classification head, training with patch size $96^3$ (sliding-window inference to target size $160^3$), baseline-default optimizer/schedule and task loss, $150$ epochs at $50$ steps/epoch, five-fold training with the last checkpoint and an ensemble of the two best folds via mean-pooled sliding-window inference.  \textbf{Segmentation.} nnU-Net-style segmentation decoder with DiceCELoss, patch size $160^3$, $150$ epochs at $100$ steps/epoch with usual nnU-Net augmentations, selecting the last checkpoint and using the best fold with nnU-Net default sliding-window inference and resampling. \textbf{Regression.} Single linear regression head with a stated post-head transform (sigmoid output scaled by $100$) followed by the baseline regression loss, training with patch size $128^3$, effective batch size $32$ using $8\times$ batch accumulation, baseline-default optimizer/schedule, $150$ epochs at $50$ steps/epoch, selected the minimum-Mean AE checkpoint, and ensembling the two best folds via Gaussian-pooled sliding-window inference with averaging. 

\subsubsection{\teamtagopen{tcFOMO2JOMO}FOMO2JOMO}

\textbf{Pretraining.} Used the pretrained BrainAgeNeXt model~\cite{LaRosa2025}, a CNN based on the MedNeXt architecture~\cite{Roy2023}. Dataset is 11{,}574 T1-weighted MRI scans from 8{,}838 subjects across 33 datasets, with metadata including age (5-95 years), scanner field strength (1.5T/3T/7T), and preprocessing steps (skull stripping, MNI152 affine registration, N4 bias-field correction).

\textbf{Finetuning.} Finetuning used one RTX A6000.
\textbf{Classification.} Decoder was a two-layer MLP classifier with dropout and a sigmoid output, used mean-across-channels stem transfer for multi-modal inputs (supporting 3-4 channels), applied focal loss, trains with Adam using differential learning rates (encoder at $0.1\times$ base, head at $1.0\times$ base). Patch size was $160\times192\times160$. Checkpoints selected based on validation accuracy. 5-fold ensemble with median probability aggregation. \textbf{Segmentation.} Used a standard MedNeXt encoder-decoder with skip connections and deep supervision disabled, adapts to two-channel inputs (DWI+FLAIR) by repeating and normalizing pretrained single-channel stem weights with optional selective freezing/differential treatment of named encoder components, optimized a combined loss of weighted cross-entropy + focal + Tversky/nDSC with automatic class-weight computation, trained with AdamW with warmup, early stopping with best-model selection by validation Dice, $160^3$ patch size. Applied spatial/intensity/noise/smoothing augmentations and channel-wise normalization, performs post-processing via largest-component retention, 3D hole filling, and nearest-neighbor interpolation to original resolution, and reports a 5-fold ensemble with probability averaging or majority vote aggregation. \textbf{Regression.} Kept the encoder and original BrainAgeNeXt regression head unchanged, states that because inputs remain T1-weighted, no weight adaptation is needed. Reported that minimal few-shot finetuning under a 5-fold cross-validation strategy did not improve over pretrained weights and therefore the model is used as originally trained, and applied a post-prediction uniform bias correction by estimating and subtracting a common bias term across subjects.

\subsubsection{\teamtagopen{tcDKFZ}DKFZ\_MIC} 
\textbf{Pretraining.} ResEnc-L CNN with 102M parameters pretrained on the OpenMind dataset~\cite{Wald2025b} using a Masked Autoencoder (MAE) objective with MSE loss. Sequences sampled i.i.d. nnUNet spatial augmentations. No hyperparameter tuning was performed beyond the default \texttt{nnssl} setup~\cite{wald2025}. Training used $160^3$ patches, batch size 8, SGD with polynomial decay over 1000 epochs / 250 steps per epoch on four A100 40GB GPUs for 46 hours.

\textbf{Finetuning.} Across all tasks when the number of input channels at fine-tuning exceeds that at pretraining, weights from the first encoder layer are repeated accordingly.
\textbf{Classification.} Classification head is linear layer on encoder features. Training used $160^3$ patches, batch size 1 with 48-step gradient accumulation, AdamW with cosine annealing for 400 epochs / 5 steps per epoch on one A100 40GB. Extensive spatial and intensity augmentations are applied alongside 0.2 label smoothing. Five-fold cross-validation with best-validation-loss checkpoint selection; test-time mirroring (8 iterations) with class-probability ensembling over the top-3 cross-validation models.
\textbf{Segmentation.} Standard U-Net decoder trained with Dice + Cross-Entropy loss. Training used $160^3$ patches, batch size 2, SGD with polynomial decay for 150 epochs on one A100 40GB, using default nnUNet augmentations and postprocessing with five-fold cross-validation checkpoint ensemble.
\textbf{Regression.} Head is a linear layer, combining pooled features from multiple scales. Trained with MSE loss using T1 scans only. Training used $160^3$ patches, batch size 4 with 48-step gradient accumulation, SGD with cosine annealing for 1000 epochs on one A100 40GB. Regularization includes 0.5 dropout in residual connections, mixup, label smoothing, shakedrop, undecay, zero-initialized residuals, and squeeze excitation, with deep supervision. Test-time mirroring (8 iterations) with age-prediction averaging over a five-model cross-validation ensemble.

\subsubsection{\teamtagopen{tcMIPLAB}MIPLAB}
\textbf{Pretraining.} U-Net (20M parameters) trained with MAE (MSE loss). Trained on FOMO60K, with curation excluding SWI, T2*, and `\texttt{scan\_*}` sequences; all remaining scans skull-stripped with volumes discarded on failure or quality grounds, yielding $48{,}000$ scans across $10{,}053$ individuals, treated i.i.d. Standard augmentations, $96^3$ patches, batch size $8$ per GPU with gradient accumulation, AdamW with cosine annealing. Trained over 12 epochs (50 steps/epoch) on 4$\times$H100 for $\sim$40 hours; last checkpoint selected. Hyperparameters tuned via grid search over patch size, LR, and batch size.

\textbf{Finetuning.} All tasks used patch size $96^3$, batch size $2$, AdamW, base LR $10^{-4}$ with cosine annealing, discriminative learning rates per encoder depth group (decay factor $2.5$), and extensive \texttt{batchgenerators}~\cite{Isensee2021} augmentation (elastic deformation, rotation/scaling, flipping, Gaussian noise/blur, brightness, gamma) applied simultaneously across all input modalities. 
\textbf{Classification.} Used organizer's \texttt{ClsRegHead}; cross-entropy loss; 21 epochs / 231 steps; $\sim$20 minutes per sweep; early stopping on \texttt{train/loss}. Bayesian HP search over LR, layer-wise learning rate decay, and loss weighting. Trained on one RTX~3090 (24GB).
\textbf{Segmentation.} Standard U-Net Decoder. DiceCE loss, trained for 31 epochs with early stopping on validation loss, progressive encoder unfreezing from deepest blocks (at epoch 5) to input convolution (epoch 25); separate encoder/decoder LRs ($3\times10^{-5}$ / $3\times10^{-4}$) under a shared cosine schedule with newly unfrozen layers inheriting the current encoder LR; foreground oversampling probability 0.85; predictions restricted to within-brain-mask voxels. Trained on one V100 (16GB).
\textbf{Regression.} Organizer's \texttt{RegHead} extended with additional linear layers, batch normalization, and dropout; modified MSE loss with variance penalty to prevent mean-age collapse; age range split into 3 equally resampled bins; gradual unfreezing/refreezing strategy over $\sim$600 epochs for $\sim$2.5 hours; early stopping; linear regression post-processing for age-level bias correction. Trained on one RTX~3090 (24GB).

\subsubsection{\teamtagopen{tcHSEE}FOMO\_HSEE}
\textbf{Pretraining.} 3D CNN, STU-Net-B~\cite{Huang2023} with 58M parameters, trained supervised on the TotalSegmentator dataset~\cite{Wasserthal2023}, containing 1143 scans (616 MRI, 527 CT).

\textbf{Finetuning.} 
\textbf{Classification.} Used a $1\times1\times1$ modality-mixing convolution to fuse four inputs (DWI, FLAIR, ADC, SWI) to a single channel plus a three-tier decoder with either log-sum-exp pooling or gated-attention MIL heads, trained with AdamW for $50$ epochs at fixed learning rates. Used BCE (class-balanced), early stopping/checkpointing, eight-flip TTA, and a $3$-fold stratified CV ensemble repeated three times. \textbf{Segmentation.} default STU-Net decoder and deep supervision, trained for $500$ epochs with SGD and nnU-Net-style polynomial decay, optimizing Dice+CE, selecting by best validation Dice, and ensembling five CV folds while excluding models with validation Dice $0$; \textbf{Regression.} Used a Regression head (gated-attention MIL) plus a modality mixer adapting inputs (noted as two channels, T1 and T2), trained for $120$ epochs with AdamW using SmoothL1 loss, early stopping eight-flip TTA, and a ten-model ensemble combining T2-only and mixed-modality variants.

\subsubsection{\teamtagopen{tcRaidium}Curia (Raidium)}
\textbf{Pretraining.} 2D Vision Transformer (ViT-B, 86M parameters) using DINOv2 \cite{dancette2025curia}. Dataset was a private, anonymized routine clinical dataset comprising 130TB and 228M DICOM files (164M CT, 64M MR), filtered to retain only 3D CT/MR exams with at least five images while removing low-quality localizer/scout sequences, without further quality control; random cropping was used as augmentation, with patch size 16, batch size 24 per GPU, using 16 A100 GPUs with mixed precision, ADAM using a reciprocal square-root learning-rate schedule for 475,000 training steps. 

\textbf{Finetuning.} For all tasks, the encoder was kept frozen. Used 1 RTX 4090. 
\textbf{Classification.} Linear decoder trained with cross-entropy loss at batch size 1 for 2,100 steps on 1 RTX 4090 using a constant learning rate selected by $3\times3$ nested cross-validation, using only DWI B1000 and ADC sequences, and averaging class and patch tokens across slices to form the representation. 
\textbf{Segmentation.} Decoder consisted of stride-2, kernel-2 transposed convolutions followed by layer normalization, GELU, and a linear layer, trained with the average of Dice and cross-entropy losses for 3,760 steps with a constant learning rate chosen by $3\times3$ nested cross-validation, using only DWI B1000 sequences and applying trilinear interpolation to original resolution plus connected-component filtering for outlier removal.
\textbf{Regression.} Linear decoder trained with MSE loss at batch size 1 for 2,000 steps using a constant learning rate selected by $3\times3$ nested cross-validation. Using T1 and T2 sequences and likewise averaging tokens across slices.

\subsubsection{\teamtagopen{tcCAMI}CAMI\_SIAT}
\textbf{Pretraining.} A UniFormer~\cite{Li2023} (\texttt{uniformer\_small}) hybrid CNN-Transformer encoder (132M parameters) from BrainMVP \cite{rui2025multi}. Trained on a dataset of 16,022 MRI scans for 200 epochs with checkpoints selected by validation performance.

\textbf{Finetuning.} Used a single A6000 GPU. \textbf{Classification.} prototype-initialized CosineSoftmax classification head with learnable temperature, trained with binary CrossEntropyLoss using Adam and cosine annealing on $96^3$ patches for 150 epochs with best-checkpoint selection by validation AUROC and no ensembling. \textbf{Segmentation.} Most encoder blocks frozen, paired with a UNet-style UniSegDecoder (with a \texttt{PatchEmbed} projection for skip alignment), optimized with CEDiceLoss using Adam and cosine annealing on patch size 64 for 1000 epochs with no ensembling. \textbf{Regression.} Excluded \texttt{patch\_embed1.proj} during weight loading to accommodate potential input-channel mismatches, decoder used global pooling over the final encoder feature for prediction, and used L1-Loss with Adam trained for 1000 epochs.

\subsubsection{\teamtagopen{tcTiger}TeamTigerAtFOMO2025}
\textbf{Pretraining.} A CNN U-Net with 20M parameters using an MAE objective. The submission followed the organizers’ base-code configuration for the model class/architecture, parameterization, and objective/loss, while augmenting the pretraining set with an additional 400 scans with glioma brain tumors covering T1, T1c, T2, and T2-FLAIR sequences. Pretraining used a 24GB GPU.

\textbf{Finetuning.} Finetuning followed the organizers’ base code for architecture (encoder/decoder and any heads), loss functions, optimization and scheduling hyperparameters, augmentation/regularization, and checkpoint selection for all three tasks. 

\subsubsection{\teamtagopen{tcMcGill}McGill\_PVG}
\textbf{Pretraining.} A ResEnc-L convolutional encoder (91M parameters) using a SimCLR-style contrastive objective optimized with the NT-XENT loss trained on the OpenMind dataset~\cite{Wald2025b}. Two independently augmented views per 3D MRI scan via a random spatial crop (minimum $60^3$) followed by resize to $160^3$, random sagittal flips and x-axis rotations up to 0.785 radians (each with probability 0.5), and intensity shifting ($\pm0.5$) plus gamma-based contrast adjustment with $\gamma\sim U[0.5,1.5]$ (each with probability $0.8$), while treating all scans and sequences as i.i.d. single-channel inputs; training used an effective batch size of 120 (10 scans/GPU across 12 GPUs, gathered before loss computation) with Adam at an initial learning rate of $3\times10^{-4}$ under a CosineAnnealingLR schedule initialized for 250 epochs but run for 24 hours to 190 epochs on 12 80GB H100 GPUs across 3 nodes with mixed precision. Final checkpoint from epoch 190 was used for all downstream experiments. 

\textbf{Finetuning.} \textbf{Classification.} using the pretrained encoder unchanged with a global-average-pooling plus single fully connected classification head, cross-entropy loss, AdamW at $10^{-4}$ with no LR schedule, patch size $160\times160\times160$, batch size 4, weight decay $3\times10{^-5}$, “all” preset augmentations, an epoch defined as 100 iterations with up to 50 epochs and early stopping after 5 non-improving epochs while saving both best-metric and final checkpoints, trained on a single 40GB A100 with mixed precision for ~30 minutes. \textbf{Segmentation.} using the same encoder with an attached 3D U-Net decoder (transpose-convolution upsampling, skip-connection concatenation, and one residual convolutional block per stage) and otherwise the same setup as for classification except a learning rate of $10^{-5}$, trained on a single 40GB A100 with mixed precision for ~45 minutes; and \textbf{Regression.} using the classification-style global-average-pooling plus single fully connected regression head, and otherwise mirrored the classification task.

\end{document}